\documentclass[acmtog]{acmart}
\acmSubmissionID{2400}
\usepackage{algorithm}
\usepackage{algpseudocode}
\usepackage{booktabs} 
\usepackage{tabularx}
\usepackage{wrapfig}
\usepackage{float}
\usepackage{graphicx}
\usepackage{pifont}
\usepackage{multirow}
\usepackage{makecell}

\usepackage{amsmath,amssymb}
\usepackage{colortbl}
\usepackage[breakable]{tcolorbox}

\usepackage{xcolor}
\newcommand{\cmark}{\textcolor{green}{\ding{51}}} 
\newcommand{\xmark}{\textcolor{red}{\ding{55}}}   
\acmJournal{TOG}

\copyrightyear{2026}
\acmYear{2026}
\setcopyright{cc}
\setcctype{by}
\acmConference[SA Conference Papers '26]{SIGGRAPH Asia 2026 Conference Papers}{December 01--04, 2026}{Kuala Lumpur, Malaysia}
\acmBooktitle{SIGGRAPH Asia 2026 Conference Papers (SA Conference Papers '26), December 01--04, 2026, Kuala Lumpur, Malaysia}
\acmDOI{10.1145/3829340.3842352}
\acmISBN{979-8-4007-2842-6/2026/12}

\begin{document}
\title{Learning 3D Editing without Paired Supervision via Generative Prior Distillation}

\author{Hao Wen}
\authornote{These authors contributed equally to this work.}
\orcid{0000-0002-4586-9880}
\affiliation{%
  \institution{Beihang University}
  \city{Beijing}
  \country{China}}
\email{wenhao1@buaa.edu.cn}

\author{Weibin Yun}
\authornotemark[1]
\orcid{0009-0006-6137-5887}
\affiliation{%
  \institution{Beihang University}
  \city{Beijing}
  \country{China}}
\email{21377079@buaa.edu.cn}

\author{Hongxing Fan}
\authornotemark[1]
\authornote{Corresponding authors.}
\orcid{0009-0003-4525-8777}
\affiliation{%
  \institution{Beihang University}
  \city{Beijing}
  \country{China}}
\affiliation{%
  \institution{Central University of Finance and Economics}
  \city{Beijing}
  \country{China}}
\email{fanhongxing@buaa.edu.cn}

\author{Haotian Lu}
\orcid{0009-0001-2643-2261}
\affiliation{%
  \institution{Beihang University}
  \city{Beijing}
  \country{China}}
\email{23241121@buaa.edu.cn}

\author{Rui Chen}
\orcid{0009-0009-7448-4275}
\affiliation{%
  \institution{Beihang University}
  \city{Beijing}
  \country{China}}
\email{crbelen@buaa.edu.cn}

\author{Zehuan Huang}
\authornote{Project Lead.}
\orcid{0009-0002-1883-0777}
\affiliation{%
  \institution{Beihang University}
  \city{Beijing}
  \country{China}}
\affiliation{%
  \institution{VAST}
  \city{Beijing}
  \country{China}}
\email{huangzehuan@buaa.edu.cn}

\author{Lu Sheng}
\orcid{0000-0002-8525-9163}
\authornotemark[2]
\affiliation{%
  \institution{Beihang University}
  \city{Beijing}
  \country{China}}
\affiliation{%
  \institution{Beijing Key Laboratory of Intelligent Creative Content Generation and Immersive Experience}
  \city{Beijing}
  \country{China}}
\email{lsheng@buaa.edu.cn}

\renewcommand\shortauthors{Wen et al.}

\begin{abstract}
Instruction-guided 3D editing is essential for interactive content creation, yet it faces a significant bottleneck: the severe scarcity of high-quality paired training data. Existing approaches attempt to bypass this by either relying on slow test-time optimization or training on pseudo-pairs constructed via complex pipelines, which often introduce structural drift and geometric artifacts. In this paper, we propose a novel framework that learns feed-forward 3D editing without paired 3D supervision via Generative Prior Distillation. Instead of relying on ground-truth 3D pairs, our core idea is to distill visual, semantic, and geometric knowledge from powerful foundation models directly into a 3D editing model. Specifically, through a differentiable rendering pipeline, we supervise the 3D representation using two complementary signals: a 2D visual prior from an image editing model at the main editing view, and a semantic prior from a Vision-Language Model at novel views to ensure strict instruction following and source identity preservation. Crucially, to address the geometric collapse and multi-view inconsistencies inherent in 2D projection supervision, we introduce a 3D-aware Distribution Matching regularization. Acting as a geometric prior, this term operates in the 3D latent space, constraining the edited output to remain within the manifold of realistic 3D assets defined by a pretrained image to 3D teacher model. Extensive experiments demonstrate that our method achieves superior instruction fidelity and cross-view consistency, significantly outperforming state-of-the-art baselines. Our project is available at: https://github.com/thiamine128/PriorEdit3D.
\end{abstract}

%
%


%
%


\begin{teaserfigure}
    \centering
  \includegraphics[width=0.85\textwidth]{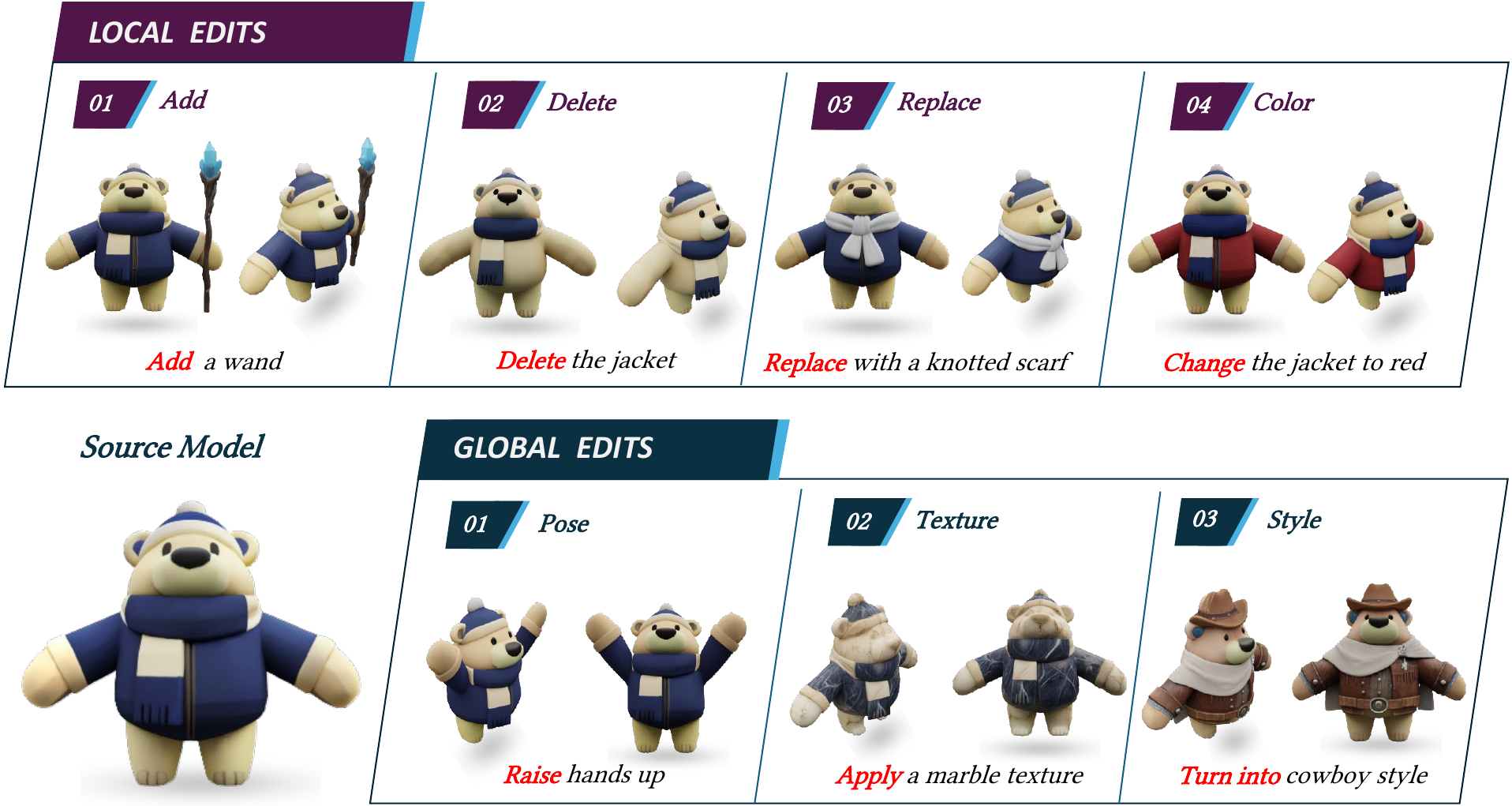}
  \caption{Our method learns feed-forward 3D editing without paired 3D supervision by distilling generative priors from foundation models, achieving high-fidelity instruction-guided editing with cross-view consistency.}
  \label{fig:teaser}
\end{teaserfigure}

\maketitle

\section{Introduction}

Large-scale 3D generative models~\cite{hong2023lrm, xiang2025structured,wen2025ouroboros3d,zhao2025hunyuan3d,wu2025direct3ds2,xiang2026native,fan20263d} have recently achieved strong geometric fidelity and visual quality. However, most of them focus on direct generation from text or image, with limited support for editable and interactive control. This limits their practicality for real-world applications that require flexible modification of existing 3D assets. Compared with 2D image editing, 3D editing is more challenging: a model must not only follow editing instructions, but also preserve geometric plausibility, maintain cross-view consistency in appearance and semantics, and retain the global structural coherence of the original asset. Crucially, training a feed-forward model to overcome these challenges typically requires massive amounts of paired 3D data (i.e., source and edited 3D assets), which is prohibitively scarce.

\begin{table}[t]
\centering
\small
\caption{\textbf{Comparison of SOTA 3D editing paradigms.} Our framework is the first unpaired approach to achieve inversion-free and mask-free editing in native 3D space. This is accomplished by distilling generative priors rather than relying on pseudo-pairs.}
\label{tab:feature_comparison}
\resizebox{\linewidth}{!}{
\renewcommand{\arraystretch}{1.3}
\begin{tabular}{@{}llccccc@{}}
\toprule
\textbf{Paradigm} & \textbf{Representative Methods} & \makecell{\textbf{Inv.-}\\\textbf{free}} & \makecell{\textbf{Native}\\\textbf{3D}} & \makecell{\textbf{No Paired}\\\textbf{3D Data}} & \makecell{\textbf{Prior}\\\textbf{Distill.}} & \makecell{\textbf{Mask-}\\\textbf{free}} \\
\midrule
Optimization & Instruct-N2N, Vox-E & \xmark & \xmark & \cmark & \cmark & \cmark \\

2D Lifting & Tailor3D, EditP23 & \cmark & \xmark & \cmark & \xmark & \cmark \\

Training-free & VoxHammer, Nano3D & \xmark & \cmark & \cmark & \xmark & \xmark \\

Supervised & 3DEditFormer, ShapeUP & \cmark & \cmark & \xmark & \xmark & \cmark \\
\midrule
\makecell[c]{Unpaired\\Distillation} & PriorEdit3D (Ours) & \textbf{\cmark} & \textbf{\cmark} & \textbf{\cmark} & \textbf{\cmark} & \textbf{\cmark} \\
\bottomrule
\end{tabular}
}
\end{table}

Due to the severe scarcity of ground-truth 3D editing pairs, early methods circumvent the need for training data by employing test-time optimization using 2D generative priors~\cite{haque2023instruct, sella2023vox}, typically through Score Distillation Sampling (SDS)~\cite{poole2022dreamfusion,Tang2023StableSD,Wang2023ProlificDreamerHA}. While SDS is effective for per-instance optimization, it is prone to view inconsistency, over-saturation, and mode collapse, and its per-scene optimization is prohibitively slow. Subsequent feed-forward approaches edit multi-view 2D images and reconstruct them into 3D models~\cite{qi2024tailor3d, baron2025editp23}. However, independent 2D modifications lack strict spatial constraints, leading to error accumulation that inevitably causes cross-view inconsistencies and geometric artifacts. To preserve geometric plausibility, recent works utilize native 3D generative priors. Training-free native methods~\cite{li2025voxhammer, ye2025nano3d} manipulate 3D latents at test time; while maintaining structural consistency, they suffer from slow latent inversion and require cumbersome manual 3D masks. 
Conversely, supervised native 3D editing models~\cite{xia2026towards, gat2026shapeup} achieve fast and mask-free inference, but they fundamentally rely on synthetic 3D pseudo-pairs. The construction of such paired data inherently restricts the diversity of editing operations (often limiting models to simple part addition, deletion, or replacement). Furthermore, these synthetic assets often exhibit structural drift, geometric distortions, and artifacts, which severely bottleneck the models' generalization and fidelity. Consequently, as summarized in table~\ref{tab:feature_comparison}, achieving scalable, fast, and robust 3D editing without paired supervision remains an unresolved challenge.

To address this problem, we present PriorEdit3D, a novel framework that formulates unpaired 3D editing as a Generative Prior Distillation process. Instead of relying on synthetic 3D pairs, our core idea is to distill visual, semantic, and geometric knowledge from powerful foundation models directly into a feed-forward 3D editing network. Specifically, at a given editing view, we differentiably render the predicted 3D representation and supervise it with a high-fidelity image generated by a 2D editing model~\cite{wu2025qwenimagetechnicalreport}, effectively distilling the 2D visual prior. To ensure the edit correctly propagates to the entire 3D space, we render the representation from novel views and introduce VLM-based semantic feedback~\cite{bai2023qwen}. In these auxiliary views, the VLM serves as a semantic prior, providing a dual-supervisory signal: (1) \textit{Instruction Following}, evaluating whether the target edit is successfully executed from unseen angles, and (2) \textit{Identity Preservation}, verifying that the edited object retains its overall semantic identity and structural coherence without unnatural distortions. Through differentiable rendering via 3D Gaussian Splatting~\cite{kerbl20233d}, both visual and semantic priors are backpropagated to optimize the 3D representation.

However, 2D distilled signals mainly constrain projections and may still cause structural drift, geometric collapse, or multi-view artifacts. To address this, we introduce 3D-aware Distribution Matching Distillation (DMD)~\cite{yin2024improved,yin2024onestep} as a geometric prior, encouraging edited outputs to stay on the pretrained 3D manifold while preserving editing semantics. We also curate a dataset and evaluation setup for unpaired 3D editing.

In summary, our main contributions are:
\begin{itemize}
    \item We propose PriorEdit3D, the first feed-forward 3D editing framework without paired 3D supervision via Generative Prior Distillation. By distilling 2D visual and VLM-based semantic priors, it achieves fast, mask-free, and highly consistent 3D manipulations.
    \item We introduce a 3D-aware distribution matching regularization that anchors edited outputs to the pretrained 3D data manifold, mitigating geometric collapse and cross-view inconsistencies from 2D-only supervision.
    \item We curate a large-scale 2D editing dataset with a rigorous evaluation protocol. Experiments show that our method produces accurate and plausible 3D edits, outperforming existing SOTA methods.
\end{itemize}

\section{Related Work}

\subsection{3D Foundation Model}

Recent advances increasingly favor native 3D generative models built on compact structured latents. 3DShape2VecSet~\cite{zhang20233dshape2vecset} compresses 3D neural fields into sets of learnable latent vectors, enabling Diffusion Transformers to model 3D data directly, while CLAY~\cite{zhang2024clay} scales this paradigm with a multi-resolution VAE ~\cite{kingma2014autoencoding} and latent DiT~\cite{peebles2023scalable} for controllable text/image-conditioned asset creation, further extending to high-resolution PBR textures via multi-view diffusion~\cite{cheng2025mvpaint,shi2023mvdream, he2025materialmvp,huang2025mv}. To achieve higher geometric fidelity at scale, recent methods combine dataset/model scaling for mesh quality (TripoSG~\cite{li2025triposg}) with sparse/hierarchical representations (Direct3D-S2~\cite{wu2025direct3ds2}; Ultra3D~\cite{chen2025ultra3d}; XCube~\cite{ren2024xcube}) and localized attention to mitigate quadratic costs on volumetric tokens. A parallel line of research focuses on unifying 3D representations and modalities through structured latents. TRELLIS~\cite{xiang2025structured} introduces a unified Structured Latent that can be seamlessly decoded into NeRF~\cite{mildenhall2021nerf}, 3D Gaussians, or meshes, mitigating cross-format incompatibilities. More recently, models further co-encode geometry and appearance/material in a single latent space: TRELLIS.2~\cite{xiang2026native} proposes O-Voxel, an omni-voxel latent jointly representing topology and PBR attributes for generating fully textured assets with efficient mesh conversion, and UniLat3D~\cite{wu2025unilat3d} unifies 3D Gaussians and meshes in a shared VAE latent space for one-stage flow-based generation.

\subsection{3D Editing}

Early methods use Score Distillation Sampling (SDS)~\cite{poole2022dreamfusion, Wang2023ProlificDreamerHA, Tang2023StableSD} to distill 2D diffusion editing priors into 3D by optimizing rendered views toward the target instruction. 
Representative works such as Instruct-NeRF2NeRF~\cite{haque2023instruct} and Vox-E~\cite{sella2023vox} apply instruction- or text-guided 2D edits to multi-view renderings and distill the edited signals back into NeRF or voxel representations via per-instance optimization. 
DreamEditor~\cite{zhuang2023dreameditor} supports localized text-driven editing of neural fields, while SketchDream~\cite{liu2024sketchdream} enables sketch- and text-guided local 3D editing. Both rely on localized per-instance optimization, whereas our method learns a reusable feed-forward editor without test-time optimization.
Recent variants further improve this optimization-based paradigm by correcting SDS gradients~\cite{alldieck2024score}, preserving identity~\cite{jin2025identitypreservingdistillationsampling}, exploiting 2D editing trajectories~\cite{wang2024enhanced3dgeneration}, enforcing cross-view correspondences~\cite{zhu2026coreeditor}, or operating in native 3D latent spaces~\cite{parelli2026latte}. 
Despite avoiding paired 3D supervision, these methods are slow and often suffer from over-saturation, mode-seeking artifacts, and cross-view inconsistency due to projection-level supervision. Vox-E~\cite{sella2023vox} further lifts 2D diffusion-based editing into a voxel representation, leveraging 3D regularization and 3D-aware attention to improve cross-view consistency. To scale beyond per-instance optimization, recent work constructs paired 3D editing data, but typically requires non-trivial pipelines and strict quality control. For example, 3DEditVerse~\cite{xia2026towards} synthesizes $\sim$118K edit pairs via separate pose-driven geometry and text-guided appearance pipelines, coupled with multi-view mask projection and filtering to reduce failures. A related lifting-based paradigm builds 3D pairs by independently reconstructing assets from 2D source/target images and curating triplets. Native 3D Editing with Full Attention~\cite{cai2025native}, built on Hunyuan3D 2.1~\cite{hunyuan3d2025hunyuan3d}, further relies on manual inspection for instruction alignment and structure preservation, yielding reliable supervision but limiting scalability. Training-free methods avoid finetuning and edit directly in native 3D representations via inversion and trajectory stabilization, including Free-Editor~\cite{karim2024free}, VoxHammer~\cite{li2025voxhammer}, and AnchorFlow~\cite{zhou2025anchorflow}. Nano3D~\cite{ye2025nano3d} leverages training-free editing to automatically bootstrap large-scale paired 3D editing datasets.

\section{Method}
In this section, we present PriorEdit3D, a feed-forward 3D editor trained {without paired 3D supervision} by distilling (i) pixel-edit loss from a strong 2D editing teacher, (ii) VLM-based semantic feedback, and (iii) DMD-based prior regularization to keep the edited results on a pretrained 3D manifold.

\begin{figure}[t]
    \centering
    \includegraphics[width=1\linewidth]{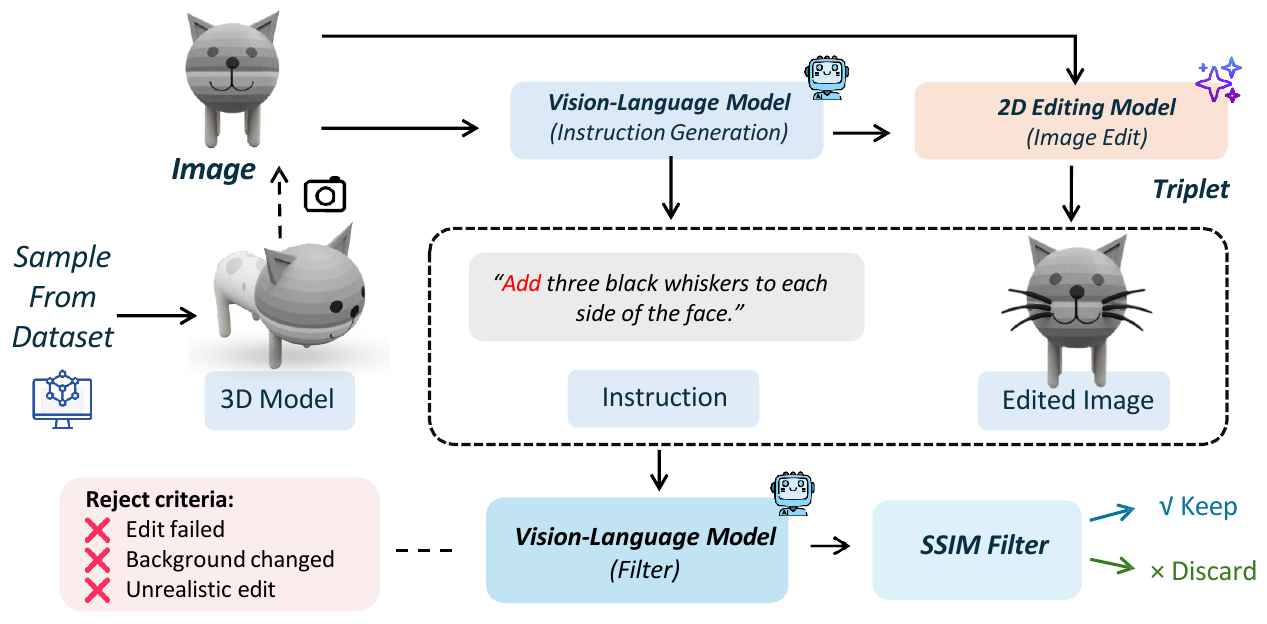}
    \caption{Automated dataset pipeline. Objaverse objects are rendered, edited via Gemini3-Flash instructions and Qwen-Image-Edit, and filtered using Qwen3-VL 32B and SSIM.}
    \label{fig:datasetpipeline}
\end{figure}

\begin{figure*}[t]
    \centering
    \includegraphics[width=1\linewidth]{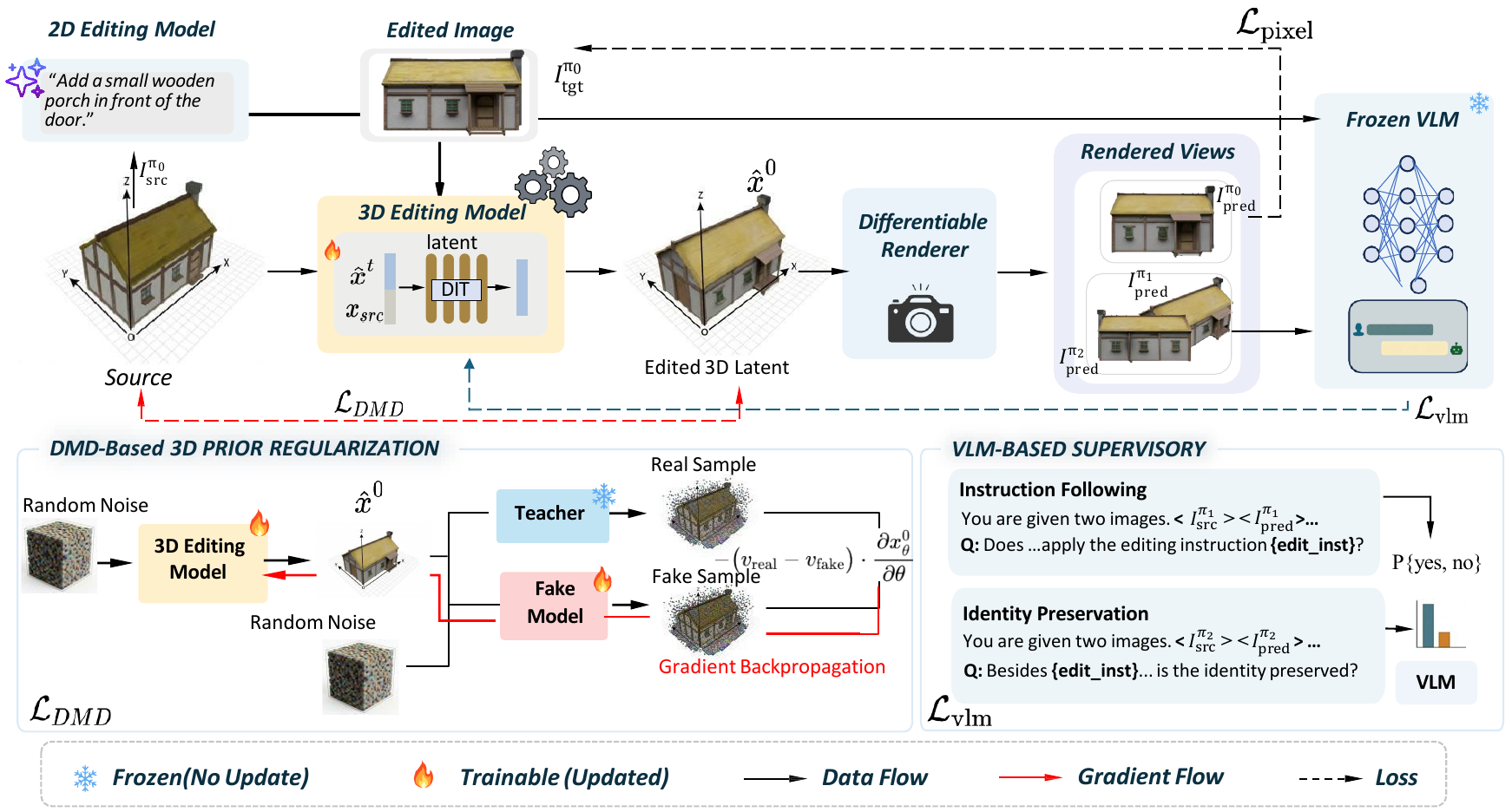}
    \caption{Training framework of PriorEdit3D. Given a source 3D latent and an instruction-conditioned edited image, the editor predicts an edited UniLat3D latent, which is rendered and optimized with pixel-level 2D supervision, VLM-based semantic/view feedback, and DMD-based 3D prior regularization.}
    \label{fig:placeholder}
\end{figure*}

\paragraph{\textnormal{\textbf{Problem formulation.}}}\quad
Let $\mathcal{A}_{\text{src}}$ be the source 3D asset with UniLat3D latent $x_{\text{src}}$~\cite{wu2025unilat3d}. Given a canonical camera pose $\pi_0$, we render the source view as
$I_{\text{src}}^{\pi_0}=\mathcal{R}(\mathcal{A}_{\text{src}},\pi_0)$.
As shown in Fig.~\ref{fig:placeholder}, a pretrained 2D editor takes $I_{\text{src}}^{\pi_0}$ and an instruction $d$ to produce the target view $I_{\text{tgt}}^{\pi_0}$, which serves as the image condition $\mathbf{c}$. Our goal is to learn a feed-forward 3D editor $G_\theta$ that predicts the edited latent $\hat{x}^{0}=G_\theta(x_{\text{src}},\mathbf{c})$. We further render two nearby novel views $\{\pi_i\}_{i=1}^{2}$ for VLM-based semantic and consistency feedback.

\paragraph{\textnormal{\textbf{Student and teacher architectures.}}}\quad
The student editor is initialized from pretrained UniLat3D. At timestep $t$, the noisy editing tokens $x^{t}\in\mathbb{R}^{B\times N\times C}$ are concatenated with the clean source tokens $x_{\mathrm{src}}\in\mathbb{R}^{B\times N\times C}$:
\begin{equation}
\widetilde{x}^{t}
=
\operatorname{Concat}_{\mathrm{token}}\!\left(x^{t},x_{\mathrm{src}}\right)
\in\mathbb{R}^{B\times 2N\times C}.
\label{eq:student_token_concat}
\end{equation}
Only $x^{t}$ is denoised, while $x_{\mathrm{src}}$ remains fixed across timesteps as source context. The student predicts $v_\theta(\widetilde{x}^{t},t,\mathbf{c})$. In contrast, the DMD teacher $G_{\mathrm{real}}$ is the frozen pretrained UniLat3D conditioned only on $\mathbf{c}$, without the source-token branch, and serves as the pretrained 3D prior during training.

\subsection{Dataset Construction}
\label{sec:dataset}
As shown in Fig.~\ref{fig:datasetpipeline}, we build a large-scale 2D editing dataset to train our unpaired 3D editing model without paired 3D supervision.

\textbf{Source Data and Instruction-Guided 2D Editing.}
We render canonical frontal views of 73,451 Objaverse objects, generate editing instructions with Gemini3-Flash~\cite{gemini3flash_eval_2025}, and apply them using Qwen-Image-Edit-2511-Lightning~\cite{modeltc_qwen_image_lightning_2026}, resulting in candidate source-edited-instruction triplets.

\textbf{Automated Curation and Quality Filtering.}
Since 2D editing models may fail to follow instructions, introduce artifacts, or modify unintended regions, we build a hybrid filtering pipeline to improve supervision reliability. We use Qwen3-VL as a multimodal judge to remove triplets with editing failures, background corruption, subject cropping, or semantic implausibility. In addition, we discard source-edited pairs with SSIM $>0.995$, which usually indicates trivial or ineffective edits.

\textbf{Taxonomy and Dataset Statistics.}
After curation, the final dataset contains 73,121 high-quality editing instances and 150,482 annotated operations, as one instance may include multiple edits, see Appendix E for more details. 
The operations are grouped into part-level and global-level edits:
\begin{itemize}
    \item \textbf{Part-level edits:} addition (46,472), removal (22,194), replacement (31,241), shape/style modification (1,774), and color modification (27,868).
    \item \textbf{Global-level edits:} pose/style/shape changes (10,744), such as pose or proportion changes and stylization, and texture changes (10,189), which modify appearance without changing geometry.
\end{itemize}

\subsection{2D Supervision via Differentiable Rendering}
\label{sec:image_loss}

Given a source latent $x_{\text{src}}$ and an editing condition $\mathbf{c}$,
we do not have paired 3D ground-truth supervision, i.e., no target 3D latent $x_{\text{tgt}}$ is available for $(x_{\text{src}},\mathbf{c})$.
Instead, we train the student 3D editor using {2D} supervision signals defined on rendered images.
Concretely, the editor generates an edited latent $\hat{x}^{0}$, which is rendered by a frozen differentiable renderer $\mathcal{R}$;
all 2D losses are then backpropagated through $\mathcal{R}$ to update the editor. We obtain $\hat{x}^{0}$ by integrating a short reverse trajectory parameterized by the student velocity field $v_\theta(\cdot)$.
We fix a decreasing timestep schedule $\{t_i\}_{i=0}^{N}$ with $t_N>\cdots>t_0=0$ and initialize $\hat{x}^{t_N}=\epsilon$, $\epsilon\sim\mathcal{N}(0,I)$.
At each iteration, we uniformly sample an exit index $s \in \{s_{\min},\ldots,s_{\max}\}$, form $\widetilde{x}^{t_i}=\operatorname{Concat}_{\mathrm{token}}(\hat{x}^{t_i},x_{\mathrm{src}})$ at every step, and integrate the reverse time dynamics using a first-order Euler discretization:
\begin{align}
\hat{x}^{t_{i-1}}
&=
\hat{x}^{t_i}
-
(t_i-t_{i-1})\, v_\theta\!\left(\widetilde{x}^{t_i},\, t_i,\, \mathbf{c}\right),
\qquad i=N,\ldots,s+1,
\label{eq:euler_prefix}\\
\hat{x}^{0}
&=
\hat{x}^{t_s}
-
t_s\, v_\theta\!\left(\widetilde{x}^{t_s},\, t_s,\, \mathbf{c}\right).
\label{eq:euler_jump}
\end{align}
To control backpropagation cost, the prefix updates in Eq.~\eqref{eq:euler_prefix} are executed with gradient stopped,
and we only backpropagate through the final jump in Eq.~\eqref{eq:euler_jump}.

\begin{figure*}[t]
    \centering
    \includegraphics[width=1\linewidth]{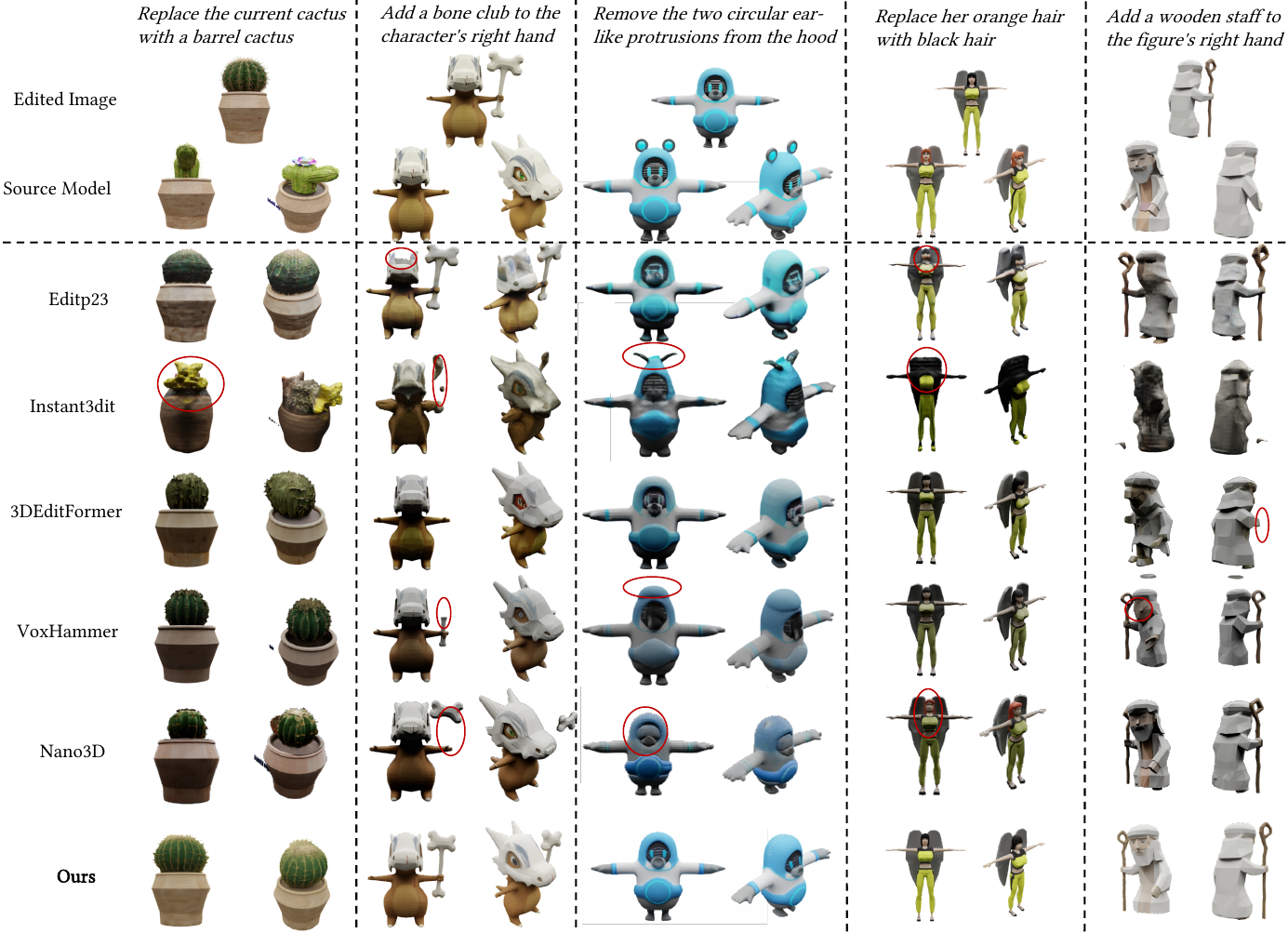}
    \caption{Qualitative comparison between our method and SOTA methods, including EditP23~\cite{baron2025editp23}, Instant3DiT~\cite{barda2025instant3dit}, 3DEditFormer~\cite{xia2026towards}, VoxHammer~\cite{li2025voxhammer}, and Nano3D~\cite{ye2025nano3d}.}
    \label{fig:qualitative}
\end{figure*}

\paragraph{\textnormal{\textbf{Pixel-level edit loss.}}}\quad
We provide dense supervision for the 3D editor on the edit view $\pi_0$ using a 2D teacher.
Specifically, the teacher supplies a view-specific edited target image $I^{\pi_0}_{\text{tgt}}$ that represents the desired edited appearance under $\pi_0$.
Given the edited latent $\hat{x}^{0}$, a differentiable 3DGS renderer $\mathcal{R}$ outputs the rendered RGB image and opacity map $(I^{\pi_0}_{\text{pred}}, O^{\pi_0}_{\text{pred}})=\mathcal{R}(\hat{x}^{0};\pi_0)$ with $O^{\pi_0}_{\text{pred}}\in[0,1]^{H\times W}$.
Optionally, we use a binary foreground mask $M^{\pi_0}\in\{0,1\}^{H\times W}$ (extracted via rembg~\cite{gatis_rembg}) to indicate the valid object region in the image plane. Mask is not required at inference, so our method remains mask-free in deployment.
Our pixel-level objective consists of three terms:
(i) a masked $\ell_1$ reconstruction loss that matches the teacher reference within the foreground,
(ii) an opacity regularization that suppresses spurious density outside the foreground region,
and (iii) a perceptual similarity term based on LPIPS:
\begin{equation}
\begin{aligned}
\mathcal{L}_{\text{pixel}}
=&\ \left\| M^{\pi_0}\odot\big(I^{\pi_0}_{\text{pred}}-I^{\pi_0}_{\text{tgt}}\big)\right\|_{1}
\\
&+\lambda_{\text{fg}}\left\|(1-M^{\pi_0})\odot O^{\pi_0}_{\text{pred}}\right\|_{1}
+\lambda_{\text{lpips}}\,\mathrm{LPIPS}\!\left(I^{\pi_0}_{\text{pred}}, I^{\pi_0}_{\text{tgt}}\right).
\end{aligned}
\label{eq:pixel_loss}
\end{equation}
The first term enforces pixel-wise consistency with the teacher      in the foreground.
The second term penalizes non-zero opacity outside the foreground mask, discouraging floating artifacts.
The third term improves perceptual fidelity by matching high-level visual features.

\paragraph{\textnormal{\textbf{VLM-based semantic feedback.}}}\quad
Pixel losses supervise only paired views, leaving unseen views prone to view-dependent overfitting, geometric degradation, or texture inconsistency. We therefore use a frozen VLM as a semantic judge~\cite{kumari2025npedit} with two complementary binary questions: \textbf{Instruction Following (IF)} on side view $\pi_1$ asks \emph{``Does the edited rendering follow the instruction?''}, while \textbf{Identity Preservation (IP)} on back view $\pi_2$ asks \emph{``Is the object identity preserved in this view?''}. This provides view-aware semantic supervision beyond the conditioning view, improving instruction alignment and back-side consistency.

Given the source latent $x_{\text{src}}$ and the edited latent $\hat{x}^{0}$, we render
$I_{\text{src}}^{\pi}=\mathcal{R}(x_{\text{src}};\pi)$ and $I_{\text{pred}}^{\pi}=\mathcal{R}(\hat{x}^{0};\pi)$ under $\pi\in\{\pi_1,\pi_2\}$.
We associate each task with a dedicated view–prompt pair:
$(\pi_{\text{IF}},\mathcal{P}_{\text{IF}})=(\pi_1,\mathcal{P}_{\text{IF}}(d))$ and
$(\pi_{\text{IP}},\mathcal{P}_{\text{IP}})=(\pi_2,\mathcal{P}_{\text{IP}})$,
where $\mathcal{P}_{\text{IF}}(d)$ is conditioned on the instruction $d$, and $\mathcal{P}_{\text{IP}}$ is a fixed preservation template (cf.fig.~\ref{fig:placeholder}).
We directly read the answer tokens (``Yes''/``No'') from the VLM's language head, convert their logit difference to a probability via sigmoid, and minimize the negative log-likelihood:
\begin{equation}
\begin{aligned}
(o_{task}^{\text{Yes}},o_{task}^{\text{No}})
&=
\text{VLMlogits}\!\left(
I_{\text{src}}^{\pi_{task}},\, I_{\text{pred}}^{\pi_{task}},\, \mathcal{P}_t
\right), \quad task\in\{\text{IF},\text{IP}\},\\
\mathcal{L}_{\text{vlm}}
&=
\sum_{task\in\{\text{IF},\text{IP}\}}
-\log \sigma\!\left(o_{task}^{\text{Yes}}-o_{task}^{\text{No}}\right).
\end{aligned}
\label{eq:vlm_unified}
\end{equation}

The VLM remains frozen, with gradients propagated through the image inputs and differentiable renderer to update only the student. 

\subsection{3D Distribution Matching Regularization}
\label{sec:dmd}

Single-view supervision weakly constrains 3D geometry: edits may match the target view but suffer from texture drift, shape distortion, or collapse from novel viewpoints. To mitigate this, we introduce Distribution Matching Distillation (DMD), using the original pretrained UniLat3D as the frozen 3D prior teacher $G_{\text{real}}$. Unlike the source-conditioned student in Eq.~\eqref{eq:student_token_concat}, the teacher operates on $(\hat{x}^{t},t,\mathbf{c})$ without receiving $x_{\mathrm{src}}$, where $\mathbf{c}=I_{\text{tgt}}^{\pi_0}$ follows UniLat3D's original image-conditioning interface.

Conditioned on $(x_{\text{src}}, c)$, the 3D editor produces an edited latent sample $\hat{x}^{0}$.
To stabilize training, we first perform an identity warm-up stage using only unedited data.
Specifically, we set the condition to the original image $c = I_{\text{src}}^{\pi_0}$ and enforce the identity mapping $G_\theta(x_{\text{src}}, I_{\text{src}}^{\pi_0}) = x_{\text{src}}$.
In edit stage training, we form $\hat{x}^{0}$ noised state at time $t\in(0,1]$ by

\begin{equation}
\hat{x}^{t}=(1-t)\hat{x}^{0}+t\epsilon,
\qquad  
\epsilon\sim\mathcal{N}(0,I).
\label{eq:noised_state_dmd}
\end{equation}

At the same state $(\hat{x}^{t},t,\mathbf{c})$, the frozen teacher provides a prior velocity
$v_{\text{real}}(\hat{x}^{t},t,\mathbf{c})$, which indicates the update direction that stays close
to the pretrained 3D data manifold.

Directly matching $v_{\text{real}}$ may over-contract the student distribution and cause mode collapse. Following DMD, we introduce a fake denoiser $F_\phi$ to estimate the student velocity field induced by $G_\theta$, denoted as $v_{\text{fake}}(\hat{x}^{t},t,\mathbf{c})$.

Comparing the teacher velocity and the fake velocity provides a
distribution-matching signal. The difference
$(v_{\text{real}}-v_{\text{fake}})$
combines
(i) an attraction term toward the pretrained prior and
(ii) a self-distribution term that prevents over-contraction.
Under the DMD formulation, this velocity difference corresponds to the
gradient of the KL divergence between the student distribution and the
teacher prior. The KL-gradient term becomes

\begin{align}
\nabla_{\theta} D_{\mathrm{KL}}
=
\mathbb{E}_{\epsilon,t}
\Big[
-\big(
v_{\text{real}}(\hat{x}^{t},t,\mathbf{c})
-
v_{\text{fake}}(\hat{x}^{t},t,\mathbf{c})
\big)
\cdot
\frac{\partial \hat{x}^{0}}{\partial \theta}
\Big],
\label{eq:kl_grad}
\end{align}

where $\partial \hat{x}^{0}/\partial\theta$ is obtained by backpropagating
through the differentiable generation steps.

We train $F_\phi$ using a standard flow-matching regression objective
on samples $\hat{x}^{0}$ from the current student distribution.
Under the velocity parameterization, the target velocity is
$v=\epsilon-\hat{x}^{0}$, and we minimize

\begin{equation}
\mathcal{L}_{\text{fake}}
=
\mathbb{E}_{\hat{x}^0,\epsilon,t}
\left\|
F_\phi(\hat{x}^{t},t,\mathbf{c})-v
\right\|_2^{2}.
\label{eq:aux_flow_matching}
\end{equation}

Finally, the editor is optimized with

\begin{equation}
\mathcal{L}
=
\mathcal{L}_{\text{pixel}}
+
\lambda_{\text{vlm}}\mathcal{L}_{\text{vlm}}
+
\lambda_{\text{DMD}}\mathcal{L}_{\text{DMD}}.
\label{eq:total_loss}
\end{equation}

We alternate updates between the editor $G_\theta$ and the fake model $F_\phi$
following the training schedule described in Appendix G.
\section{Experiments}
\label{sec:experiments}

\begin{table*}[h]
    \centering
    \caption{Main quantitative comparison across different evaluation datasets. We report edited-image alignment, overall 3D quality, condition alignment, and runtime per edit on a single A100 GPU.}
    \label{tab:main_results}
    \setlength{\tabcolsep}{8pt}
    \renewcommand{\arraystretch}{1.12}
    \small
    \begin{tabular}{l|ccc|cc|ccc|c}
    \toprule
    & \multicolumn{3}{c|}{\textbf{Edited Image Alignment}} 
    & \multicolumn{2}{c|}{\textbf{Overall 3D Quality}} 
    & \multicolumn{3}{c|}{\textbf{Condition Alignment}} 
    & \\
    \cmidrule{2-9}
    \textbf{Method}
    & PSNR $\uparrow$ 
    & SSIM $\uparrow$ 
    & LPIPS $\downarrow$ 
    & FID $\downarrow$ 
    & FVD $\downarrow$ 
    & CLIP-T $\uparrow$ 
    & \makecell{LLM-Id\\(\%)$\uparrow$} 
    & \makecell{LLM-Inst\\(\%)$\uparrow$}
    & \textbf{Runtime} \\
    \midrule

    \rowcolor{gray!12}
    \multicolumn{10}{c}{\textbf{PriorEdit3D}} \\
    \midrule
    EditP23~\cite{baron2025editp23}       
    & 16.74 & 0.87 & 0.19 & 184.51 & 286.72 & 0.29 & 82.52 & 64.71 & 18 s \\
    Instant3DiT~\cite{barda2025instant3dit}   
    & 15.45 & 0.86 & 0.20 & 229.47 & 328.24 & 0.23 & 40.09 & 25.22 & 20 s \\
    3DEditFormer~\cite{xia2026towards}  
    & 18.89 & 0.89 & 0.15 & 167.85 & 220.91 & 0.28 & 86.31 & 53.39 & 74 s \\
    VoxHammer~\cite{li2025voxhammer}     
    & 16.87 & 0.87 & 0.17 & 179.22 & 359.50 & 0.27 & 76.54 & 51.74 & 133 s \\
    Nano3D~\cite{ye2025nano3d}     
    & 19.90 & 0.91 & \textbf{0.07} & 103.50 & \textbf{189.93} & 0.21 & 85.35 & 60.37 & 14s \\
    \midrule
    {\bfseries Ours} 
    & \textbf{24.37} & \textbf{0.94} & 0.12 & \textbf{71.96} & 195.77 & \textbf{0.31} & \textbf{93.13} & \textbf{86.92} & \textbf{7 s} \\

    \midrule
    \rowcolor{gray!12}
    \multicolumn{10}{c}{\textbf{ABO \& GSO}} \\
    \midrule
    EditP23~\cite{baron2025editp23}       
    & 19.13 & 0.89 & \textbf{0.10} & 143.31 & 224.74 & 0.14 & 64.64 & 11.90 & \cellcolor{gray!15}-- \\
    3DEditFormer~\cite{xia2026towards}  
    & 17.31 & 0.88 & 0.15 & 127.54 & 197.22 & 0.18 & 83.99 & 67.86 & \cellcolor{gray!15}-- \\
    Nano3D~\cite{ye2025nano3d}     
    & 18.39 & 0.87 & \textbf{0.10} & 121.66 & 270.86 & 0.20 & 83.50 & 63.26 & \cellcolor{gray!15}-- \\
    \midrule
    {\bfseries Ours} 
    & \textbf{20.59} & \textbf{0.89} & 0.13 & \textbf{117.11} & \textbf{168.78} & \textbf{0.23} & \textbf{91.14} & \textbf{68.41} & \cellcolor{gray!15}-- \\
    \bottomrule
    \end{tabular}
\end{table*}

\begin{figure*}[t]
    \centering
    \includegraphics[width=0.9\linewidth]{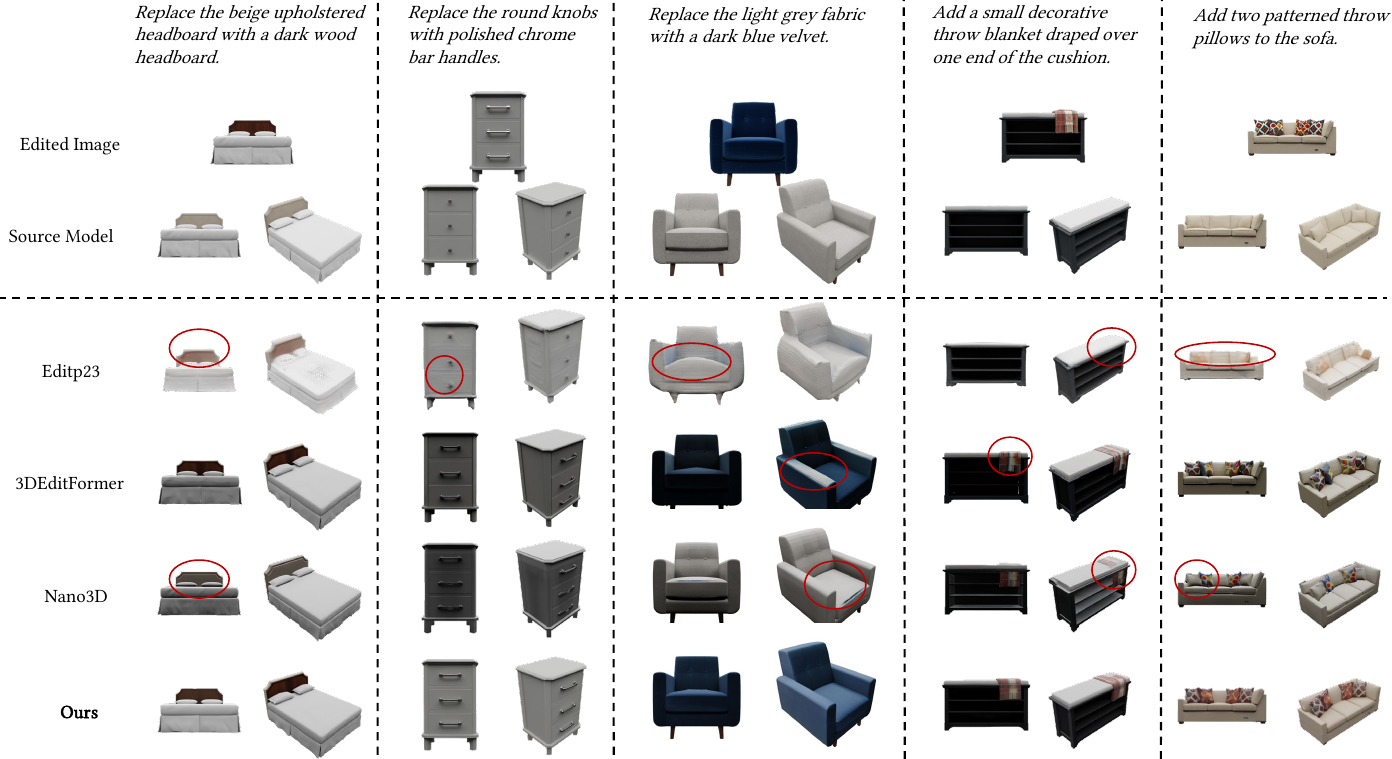}
    \caption{\textbf{Qualitative results on ABO dataset.}}
    \label{fig:result_abo}
\end{figure*}

\begin{figure*}[t]
    \centering
    \includegraphics[width=0.9\linewidth]{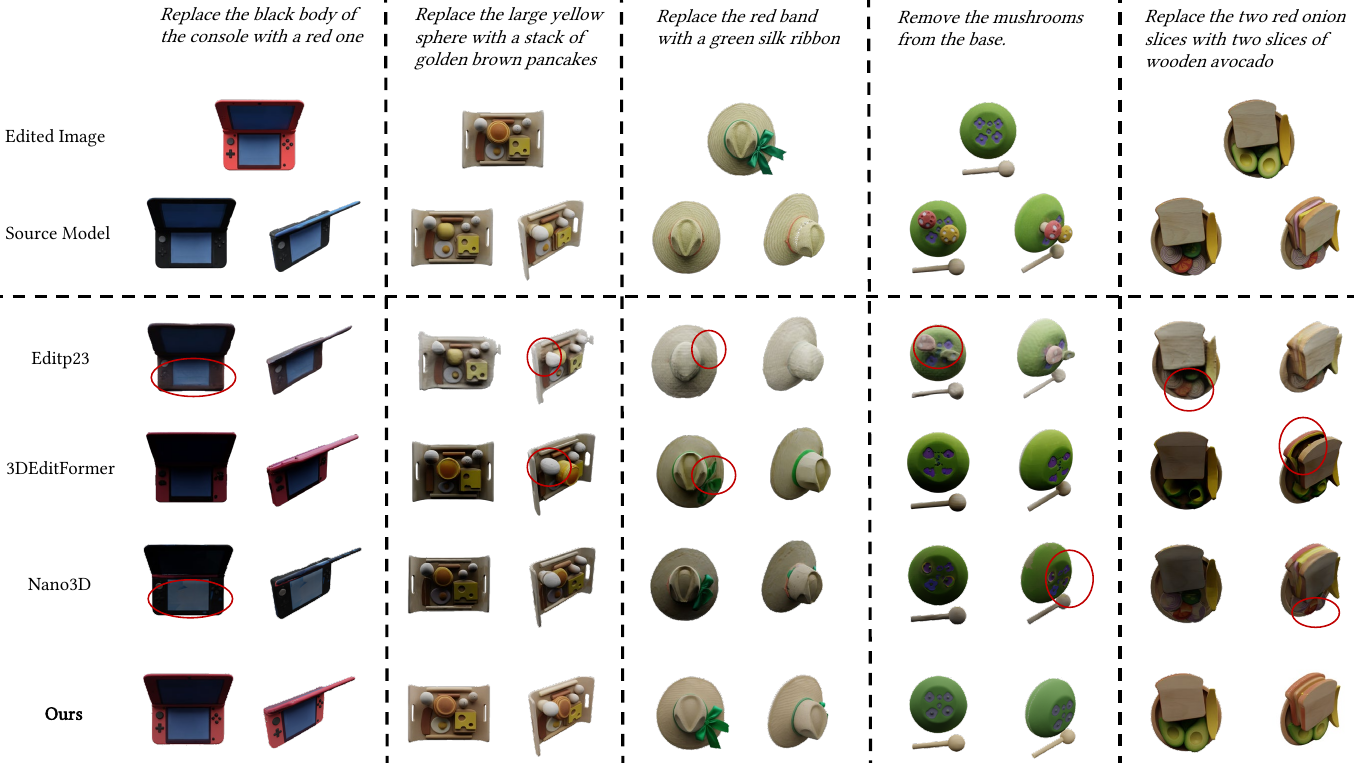}
    \caption{\textbf{Qualitative results on GSO dataset.}}
    \label{fig:result_gso}
\end{figure*}

\subsection{Experimental Setup}
\label{sec:qualitative}

\paragraph{\textnormal{\textbf{Datasets}}}

We train our model using the curated dataset described in section ~\ref{sec:dataset}. For in-distribution evaluation, we construct a held-out test set of 130 representative samples, balanced across 3D assets and editing types. To assess out-of-distribution generalization, we further build test sets from Amazon Berkeley Objects (ABO) and Google Scanned Objects (GSO) with manually constructed instructions. In total, our evaluation includes 130 curated samples, 236 ABO samples, and 239 GSO samples.

\paragraph{\textnormal{\textbf{Implementation Details}}}
We build on UniLat3D with Qwen3-VL-4B as the frozen VLM feedback model, and adapt the generator into an editor via token-concatenation conditioning. Training uses AdamW ($lr=10^{-5}, \beta_1=0.0, \beta_2=0.9$) for 18K iterations on 8 A100 GPUs, with 10 fake-model updates per editor step. Inference takes 20 sampling steps on a single A100 GPU.

\paragraph{\textnormal{\textbf{Evaluation}}}
For fair comparison, all 3D assets are aligned to a unified coordinate system and rendered with Blender Cycles at $512 \times 512$ resolution. For image-based metrics, we use three fixed views: front, back, and an angled conditional view. To evaluate multi-view consistency, we additionally render 24-frame $360^\circ$ turntable videos at 8 FPS. For methods that require 3D editing masks (VoxHammer and Instant3DiT), we manually annotate 3D editing regions on our dataset to ensure a fair evaluation.

\paragraph{\textnormal{\textbf{Metrics}}}
\label{sec:metrics}

We evaluate three aspects: \emph{edited-image alignment} using PSNR, SSIM, and LPIPS between the 2D edited image and the rendered 3D output from the same view; \emph{overall 3D quality} using FID~\cite{heusel2017gans} and FVD~\cite{unterthiner2018towards} on multi-view renderings; and \emph{condition alignment} using CLIP-T~\cite{radford2021learning} and Qwen2.5-VL-72B-Instruct~\cite{Qwen2.5-VL} scores for identity preservation (LLM-Id, \%) and instruction following (LLM-Inst, \%). Details are in 
Appendix A.

\subsection{Main Results}
As shown in Fig.~\ref{fig:qualitative}, Fig.~\ref{fig:result_abo} and Fig.~\ref{fig:result_gso}, we compare PriorEdit3D with EditP23, Instant3DiT, 3DEditFormer, and VoxHammer across diverse editing scenarios. The qualitative results align with Table~\ref{tab:main_results}: PriorEdit3D achieves the best overall performance, with higher edited-image alignment, better 3D quality, stronger condition alignment, and improved identity preservation.

Existing methods struggle to balance editing fidelity and structural consistency. Lifting-based EditP23 often over-smooths geometry and removes unedited details, such as the character's face in the bone club case. Instant3DiT produces severe artifacts and unstable topology, as seen in the corrupted barrel cactus and collapsed blocky hair. Native 3D editors show different limitations. 3DEditFormer preserves source identity but overly constrains new geometry, failing to generate the barrel cactus or add the wooden staff. VoxHammer preserves unedited regions through latent replacement, but relies on precise manual 3D masks and rigid spatial anchoring. As a result, it suffers from boundary artifacts and clipping when edits change the footprint, such as the incomplete bone club and clipped wooden staff, and cannot handle global transformations such as pose or major shape changes. Nano3D maintains relatively stable global geometry, but its edits are often incomplete or weakly localized, causing poor object attachment, residual artifacts after part removal, and partial appearance changes such as incomplete hair recoloring.

In contrast, our method achieves precise localized edits while preserving global geometry, multi-view consistency, and unedited regions, yielding fewer artifacts, better structural/texture consistency, stronger instruction following, and less identity drift. As shown in Table~\ref{tab:main_results}, PriorEdit3D edits a 3D asset in 7 seconds via an inversion-free feed-forward pipeline, much faster than the baselines.

\begin{figure*}[t]
    \centering
    \includegraphics[width=0.9\linewidth]{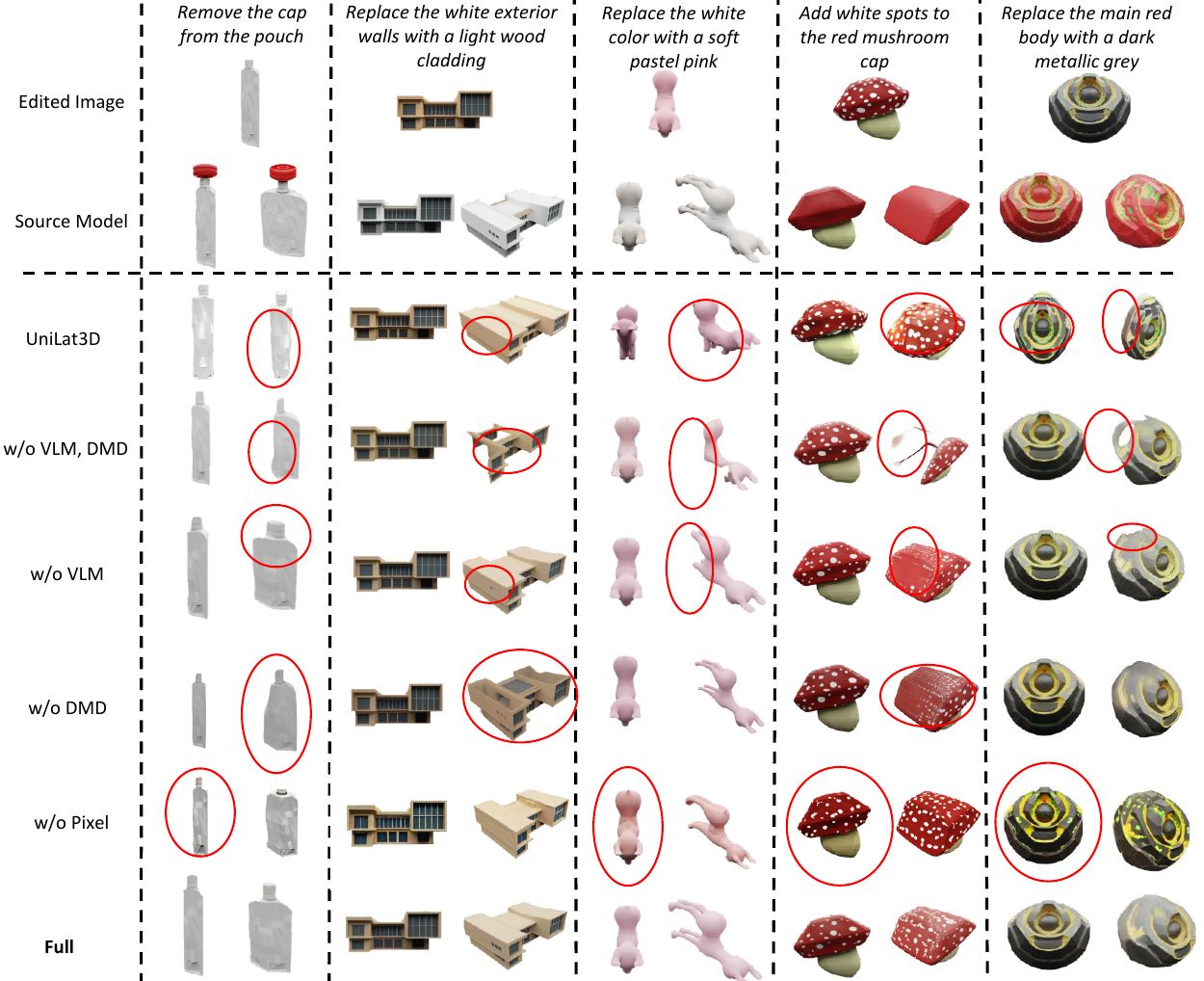}
    \caption{\textbf{Qualitative ablation results.} We compare our full method with variants w/o VLM, w/o DMD, w/o Pixel, w/o both VLM and DMD, and direct conditional generation (UniLat3D). Red circles highlight failures: removing VLM degrades non-edited views; removing DMD causes geometric inconsistencies; removing pixel supervision weakens edit fidelity; removing both VLM and DMD causes severe collapse; and UniLat3D fails to preserve source geometry. Our full model ensures accurate edits and 3D consistency.}
    \label{fig:ablation}
\end{figure*}

\subsection{Ablation Study}
\label{sec:ablation_components}

\subsubsection{Impact of Loss Term}

Table~\ref{tab:ablation_loss} and Fig.~\ref{fig:ablation} show that our losses are complementary. Pixel supervision anchors conditioning-view fidelity and source identity, DMD improves 3D realism and cross-view consistency, and VLM feedback strengthens semantic alignment in novel views. The leave-one-out variants further show that VLM guidance cannot replace dense pixel supervision. Raw UniLat3D preserves texture consistency but fails to retain source geometry.

\begin{table}[t]
    \centering
    \caption{Ablation study of loss terms. We compare pixel reconstruction loss, VLM loss, DMD loss, and additional leave-one-out ablations.}
    \label{tab:ablation_loss}
    \setlength{\tabcolsep}{6pt}
    \renewcommand{\arraystretch}{1.12}
    \resizebox{\linewidth}{!}{
    \begin{tabular}{ccc|ccc|cc|ccc}
        \toprule
        \multicolumn{3}{c|}{\textbf{Loss Terms}} &
        \multicolumn{3}{c|}{\textbf{Edited Image Align}} &
        \multicolumn{2}{c|}{\textbf{Overall 3D Quality}} &
        \multicolumn{3}{c}{\textbf{Condition Alignment}} \\
        \cmidrule{1-11}
        \textbf{Pixel} & \textbf{DMD} & \textbf{VLM} &
        PSNR $\uparrow$ & SSIM $\uparrow$ & LPIPS $\downarrow$ &
        FID$\downarrow$ & FVD$\downarrow$ & CLIP-T$\uparrow$ &
        \makecell{LLM-Id\\(\%)$\uparrow$} &
        \makecell{LLM-Inst\\(\%)$\uparrow$} \\
        \midrule

        \cmark & \xmark & \xmark &
        24.00 & 0.94 & 0.12 & 78.90 & 307.58 & 0.30 & 88.35 & 80.38 \\

        \cmark & \xmark & \cmark &
        23.21 & 0.93 & 0.12 &
        75.13 & 211.08 & \textbf{0.31} &
        90.10 & \textbf{86.92} \\

        \cmark & \cmark & \xmark &
        \textbf{24.45} & 0.94 & 0.12 &
        73.67 & \textbf{193.44} & 0.30 & 90.92 & 83.46 \\

        \xmark & \cmark & \cmark &
        20.99 & 0.91 & 0.14 &
        94.11 & 197.79 & 0.30 &
        88.29 & \textbf{86.92} \\

        \cmark & \cmark & \cmark &
        24.37 & 0.94 & 0.12 &
        \textbf{71.96} & 195.77 & \textbf{0.31} &
        \textbf{93.13} & \textbf{86.92} \\
        \bottomrule
    \end{tabular}
    }
\end{table}

\subsubsection{Impact of VLM Backbone}
\label{sec:supp_vlm_backbone}
We study the effect of different VLM backbones used for semantic feedback during training. All variants share the same prompt template and training configuration; only the VLM backbone is varied. As shown in Table~\ref{tab:vlm_type_ablation}, the 4B VLM slightly outperforms the 2B variant, suggesting more accurate semantic feedback.

\begin{table}[h]
    \centering
    \caption{Impact of VLM backbone on 3D editing performance.}
    \label{tab:vlm_type_ablation}
    \setlength{\tabcolsep}{6pt}
    \renewcommand{\arraystretch}{1.12}
    \resizebox{\linewidth}{!}{
        \begin{tabular}{l|ccc|cc|ccc}
        \toprule
        \multirow{2}{*}{\textbf{VLM Backbone}} &
        \multicolumn{3}{c|}{\textbf{Edited Image Align}} &
        \multicolumn{2}{c|}{\textbf{Overall 3D Quality}} &
        \multicolumn{3}{c}{\textbf{Condition Alignment}} \\
        \cmidrule{2-9}
        & PSNR (M)$\uparrow$ & SSIM (M)$\uparrow$ & LPIPS (M)$\downarrow$ &
          FID$\downarrow$ & FVD$\downarrow$ & CLIP-T$\uparrow$ & \makecell{LLM-Id\\(0--100)$\uparrow$} & \makecell{LLM-Inst\\(\%)$\uparrow$} \\
        \midrule
        Qwen3-VL-2B  & 24.26 & 0.94 & 0.12 & 74.34 & 195.90 & 0.30 & 91.50 & 85.77 \\
        Qwen3-VL-4B  & \textbf{24.37} & \textbf{0.94} & 0.12 & 71.96 & 195.77 & \textbf{0.31} & \textbf{93.13} & \textbf{86.92} \\
        \bottomrule
        \end{tabular}
    }
\end{table}

\subsubsection{Impact of VLM Prompt Feedback}
\label{sec:supp_prompt_feedback}
We ablate the VLM prompt used for semantic feedback during training. All variants use the same VLM backbone and training setup, varying only the prompt strategy.
As Table ~\ref{tab:vlm_prompt_ablation} shows, with only the edit fidelity prompt or the identity prompt, they perform worse than the full model in terms of condition alignment. While using the identity-only prompt can improve overall 3D quality, it performs slightly worse in instruction following compared to the pure edit fidelity prompt.

\begin{table}[h]
    \centering
    \caption{Impact of VLM prompt strategy on 3D editing performance.}
    \label{tab:vlm_prompt_ablation}
    \setlength{\tabcolsep}{6pt}
    \renewcommand{\arraystretch}{1.12}
    \resizebox{\linewidth}{!}{
        \begin{tabular}{l|ccc|cc|ccc}
        \toprule
        \multirow{2}{*}{\textbf{Prompt Strategy}} &
        \multicolumn{3}{c|}{\textbf{Edited Image Align}} &
        \multicolumn{2}{c|}{\textbf{Overall 3D Quality}} &
        \multicolumn{3}{c}{\textbf{Condition Alignment}} \\
        \cmidrule{2-9}
        & PSNR (M)$\uparrow$ & SSIM (M)$\uparrow$ & LPIPS (M)$\downarrow$ &
          FID$\downarrow$ & FVD$\downarrow$ & CLIP-T$\uparrow$ & \makecell{LLM-Id\\(0--100)$\uparrow$} & \makecell{LLM-Inst\\(\%)$\uparrow$} \\
        \midrule
        Edit fidelity only                   & 23.10 & 0.93 & 0.12 & 80.28 & 211.90 & \textbf{0.31} & 90.21 & 83.85 \\
        Identity only                    & 24.33 & \textbf{0.94} & 0.12 & \textbf{71.77} & \textbf{187.01} & 0.30 & 91.94 & 82.31 \\
        Full                  & \textbf{24.37} & \textbf{0.94} & 0.12 & 71.96 & 195.77 & \textbf{0.31} & \textbf{93.13} & \textbf{86.92} \\
        \bottomrule
        \end{tabular}
    }
\end{table}

\subsubsection{VLM Metric Backbone}

\begin{figure*}[t]
    \centering
    \begin{minipage}{0.52\linewidth}
        \centering
        \includegraphics[width=\linewidth]{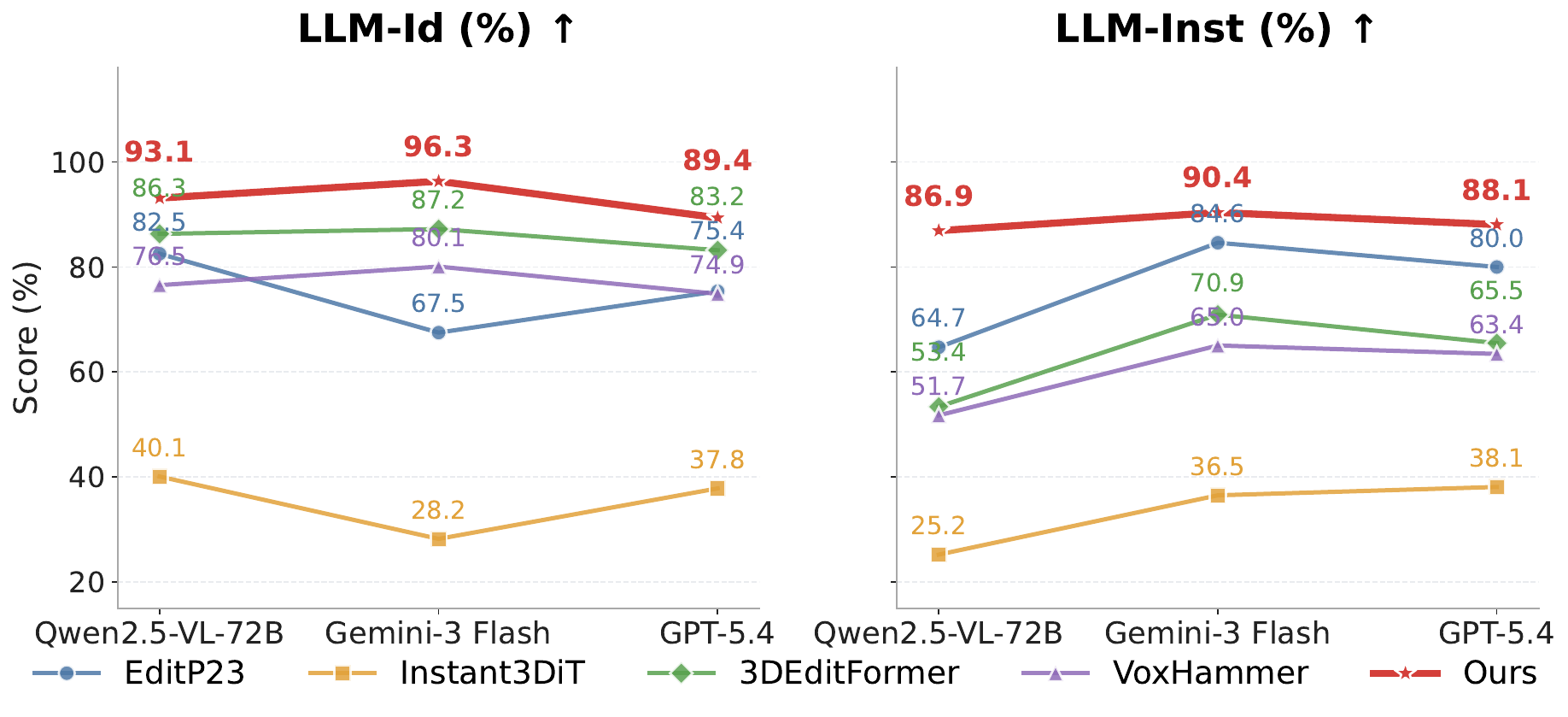}
        \caption{VLM Evaluation with different judge backbones: Qwen2.5-VL-72B, Gemini-3 Flash, and ChatGPT 5.4.}
        \label{fig:supp_vlm_eval}
    \end{minipage}
    \hfill
    \begin{minipage}{0.45\linewidth}
        \centering
        \includegraphics[width=\linewidth]{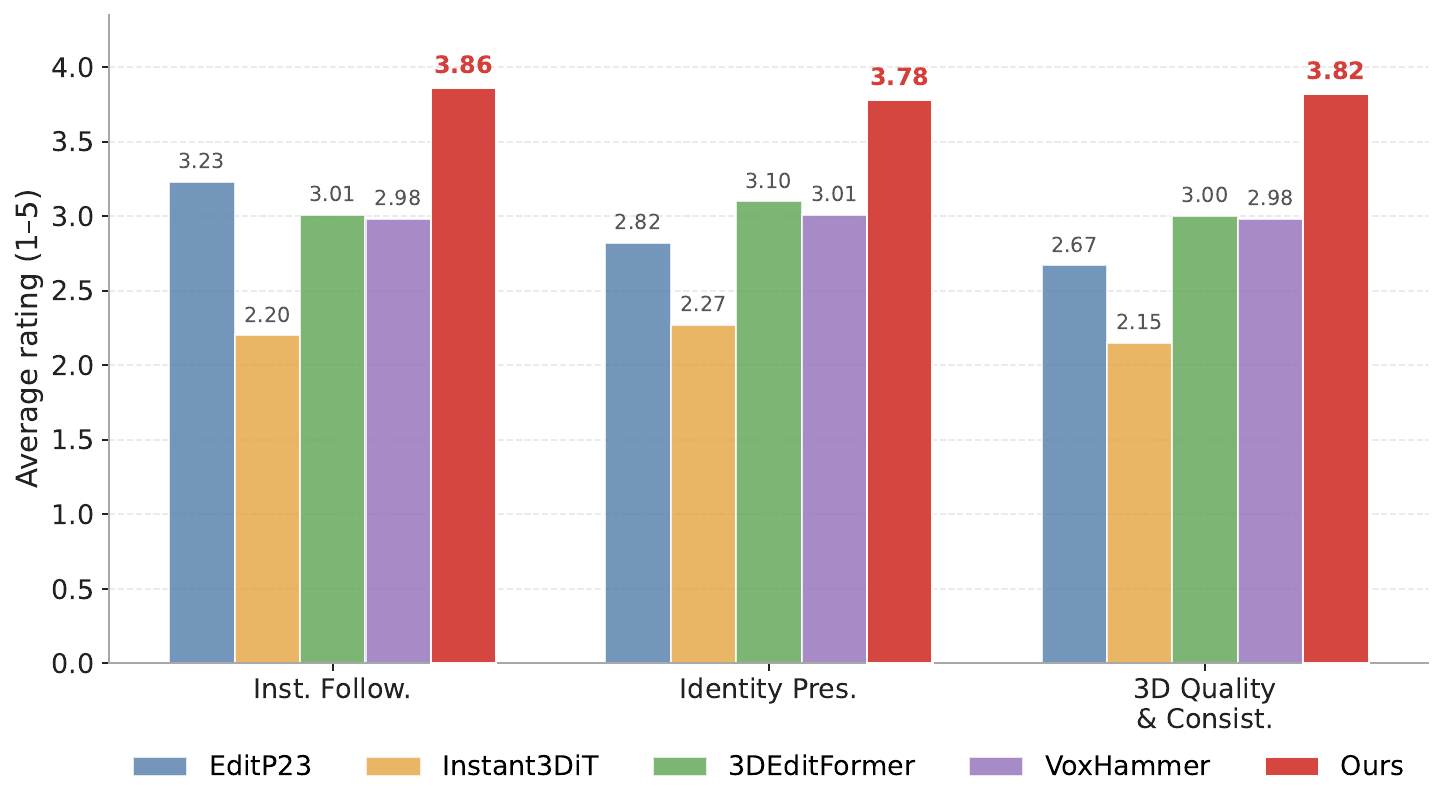}
        \caption{\textbf{User Study Results.} Average human ratings on a 1-5 scale.}
        \label{fig:user_study}
    \end{minipage}
\end{figure*}

With Qwen2.5-VL-72B as the default evaluator, we also use Gemini3-Flash and GPT-5.4 to re-compute LLM-Id and LLM-Inst. Fig.~\ref{fig:supp_vlm_eval} shows that our method consistently ranks first across all three VLM judges.

\begin{figure*}[t]
    \centering
    \includegraphics[width=0.9\linewidth]{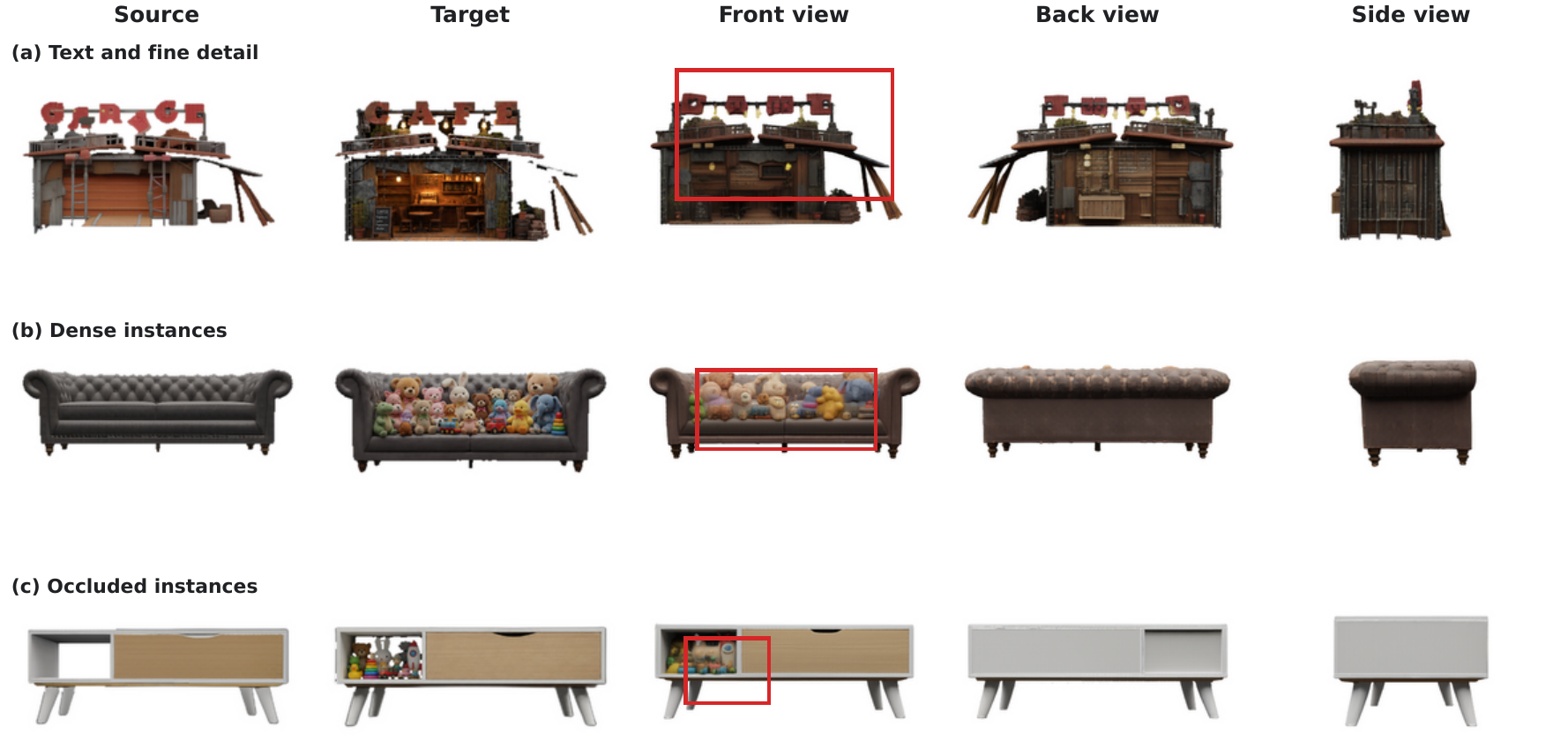}
    \caption{Representative failure cases.}
    \label{fig:failure}
\end{figure*}

\subsection{User Study}

We conduct a user study against four SOTA baselines. Each of the 50 participants evaluates 20 randomly sampled editing scenarios. In each trial, participants view the source 360$^\circ$ video, text instruction, 2D edited image, and anonymized results from all five methods in randomized order. They rate each result on a 1--5 Likert scale for Instruction Following, Identity Preservation, and 3D Quality and Consistency. As shown in Fig.~\ref{fig:user_study}, our method achieves the highest ratings across all criteria. It outperforms mask-based VoxHammer in Identity Preservation and 3D Quality, demonstrating the effectiveness of prior distillation and 3D distribution matching regularization.

\subsection{Limitation}
\label{sec:limitation}
While effective, our method has several limitations. As shown in Fig. \ref{fig:failure}, it struggles with fine-grained edits such as precise text and dense or small instances, as global consistency is prioritized over high-frequency details. Non-edited regions may still exhibit unintended texture or geometric drift. Errors from the pretrained 2D editor may also propagate to the reconstructed 3D model. Finally, the method is constrained by the UniLat3D prior and differentiable rendering, making large pose or topology changes difficult and limiting compatibility with non-differentiable 3D architectures.

\section{Conclusion}
\label{sec:conclusion}

In this work, we present a novel framework PriorEdit3D for 3D editing that eliminates the need for paired 3D supervision.
By distilling pixel-level guidance from a pretrained 2D editing model and semantic supervision from a Vision-Language Model, our method learns a feed-forward 3D editor entirely from unpaired 3D data.
To mitigate the geometric collapse and multi-view inconsistencies that arise from 2D-only supervision, we further introduce a 3D-aware Distribution Matching regularization that constrains edited outputs to lie on the manifold of realistic 3D assets.
Extensive experiments demonstrate that our approach  outperforms existing baselines in both instruction fidelity and cross-view consistency, validating the effectiveness of generative prior distillation as a scalable alternative to paired data collection for 3D editing.

\begin{acks}
This work was supported by the National Natural Science Foundation of China (62132001), the Beijing Natural Science Foundation (L252218), and the Fundamental Research Funds for the Central Universities.
\end{acks}

\bibliographystyle{ACM-Reference-Format}
\bibliography{bibliography}

\clearpage
\appendix


\section{LLM Evaluation Metric Prompts}
\label{sec:supp_llm_metric_prompts}

We detail the prompt templates used for the two VLM-based evaluation metrics introduced in Section~\ref{sec:metrics}. 

\paragraph{LLM-Id (Identity Preservation).}
To evaluate identity preservation, we independently compare the front views and the back views of the source and edited 3D assets. For each view pair (e.g., source front and edited front), we query the VLM with the following prompt and calculate the average of the two scores:

\begin{tcolorbox}[colback=gray!5, colframe=gray!60, title=LLM-Id Prompt Template]
\small
\texttt{You are evaluating visual quality of a 3D edited object. You are given:
- Image 1: Rendering of the original 3D model (before editing).
- Image 2: Rendering of the edited 3D model (after editing).
Question:
Is the object edited keeping the original model's style and preserving its identity?
IGNORE the changes in the second image because of the editing instruction `\{instruction\}'.
Provide a score from 0 to 100. Answer strictly with an integer number between 0 and 100. Do not explain.}
\end{tcolorbox}

\paragraph{LLM-Inst (Instruction Following).}
To evaluate instruction following, we independently assess the front and conditional views of the edited asset. We provide the VLM with a single rendered view at a time along with the editing instruction, querying the VLM with:

\begin{tcolorbox}[colback=gray!5, colframe=gray!60, title=LLM-Inst Prompt Template]
\small
\texttt{You are evaluating visual quality of a 3D edited object. You are given:
- Image: Rendering of the edited 3D model (after editing).
Question:
Does edited object successfully apply the editing instruction `\{instruction\}'?
Answer strictly with "Yes" or "No". Do not explain.}
\end{tcolorbox}

For the LLM-Inst metric, we map the VLM's ``Yes'' and ``No'' responses to $1$ and $0$, respectively. The final score is calculated as the average success rate across both the front and conditional views for all test assets, reported as a percentage.

\section{Distribution Matching Distillation Formulation}
\label{sec:supp_dmd}

We provide the formulation of the Distribution Matching Distillation (DMD) objective used in our framework.
Let $p_{\phi}(x)$ denote the pretrained 3D generative prior (teacher) over realistic 3D latents,
and let $q_{\theta}(x)$ denote the distribution induced by our 3D editing model (student).
Our goal is to align the edited latent distribution with the pretrained prior by minimizing the reverse KL divergence
\begin{equation}
\mathcal{L}_{\mathrm{DMD}}
=
D_{\mathrm{KL}}\!\left(q_{\theta}(x)\,\|\,p_{\phi}(x)\right).
\label{eq:dmd_kl}
\end{equation}

Although the densities of $q_{\theta}$ and $p_{\phi}$ are intractable, we only need the gradient of
Eq.~\eqref{eq:dmd_kl} with respect to $\theta$.
Following the original DMD derivation, this gradient can be written as
\begin{equation}
\nabla_{\theta}\mathcal{L}_{\mathrm{DMD}}
=
\mathbb{E}_{x\sim q_{\theta}}
\left[
\Big(
\nabla_x \log q_{\theta}(x)
-
\nabla_x \log p_{\phi}(x)
\Big)
\frac{\partial x}{\partial \theta}
\right],
\label{eq:dmd_grad_exact}
\end{equation}
which shows that optimization is driven by the difference between the score of the student distribution
and that of the teacher distribution.

Directly evaluating Eq.~\eqref{eq:dmd_grad_exact} is difficult, since diffusion models estimate scores on
\emph{perturbed} distributions rather than on the clean data distribution itself.
Following DMD, we therefore inject Gaussian noise into the student sample.
Given a clean latent $x\sim q_{\theta}$, we sample a timestep $t$ and Gaussian noise
$\epsilon\sim\mathcal{N}(0,I)$, and construct the perturbed latent
\begin{equation}
x_t = \alpha_t x + \sigma_t \epsilon,
\label{eq:dmd_noised_latent}
\end{equation}
where $\alpha_t$ and $\sigma_t$ are determined by the diffusion noise schedule.

We use two diffusion denoisers to approximate the score functions on the perturbed distributions.
The \emph{real} denoiser is a frozen copy of the pretrained teacher prior and provides the score of the perturbed teacher distribution.
The \emph{fake} denoiser is trained to model the score of the perturbed student distribution.
With a slight abuse of notation, we denote the corresponding score estimators by
$s_{\phi}(x_t,t)$ and $s_{\psi}(x_t,t)$, respectively.
The practical DMD gradient is then approximated as
\begin{equation}
\nabla_{\theta}\mathcal{L}_{\mathrm{DMD}}
\approx
\mathbb{E}_{x\sim q_{\theta},\,t,\,\epsilon}
\left[
w(t)\,
\big(
s_{\psi}(x_t,t)-s_{\phi}(x_t,t)
\big)
\frac{\partial x}{\partial \theta}
\right],
\label{eq:dmd_grad_approx}
\end{equation}
where $w(t)$ is a timestep-dependent weighting factor used to balance gradients across noise levels.

The teacher score estimator $s_{\phi}$ is fixed throughout training.
Since the student distribution $q_{\theta}$ changes as the editor is updated,
the fake diffusion model must be trained online to track the current student distribution.
Following the original DMD formulation, we optimize the fake model with a standard diffusion denoising objective:
\begin{equation}
\mathcal{L}_{\mathrm{fake}}
=
\mathbb{E}_{x\sim q_{\theta},\,t,\,\epsilon}
\left[
\lambda(t)\,
\left\|
\mu_{\psi}(x_t,t)-x
\right\|_2^2
\right],
\label{eq:dmd_fake_loss}
\end{equation}
where $\mu_{\psi}(x_t,t)$ denotes the denoised prediction of the fake diffusion model,
and $\lambda(t)$ is the same timestep weighting strategy as used in diffusion training.

In practice, training alternates between two updates:
(i) updating the editor parameters $\theta$ using the score-difference signal in
Eq.~\eqref{eq:dmd_grad_approx}, and
(ii) updating the fake diffusion model parameters $\psi$ using Eq.~\eqref{eq:dmd_fake_loss}.
This procedure encourages the edited latent distribution to move toward the pretrained 3D prior
while continuously correcting for the evolving student distribution.

\begin{figure*}[t]
    \centering
    \includegraphics[width=0.85\linewidth]{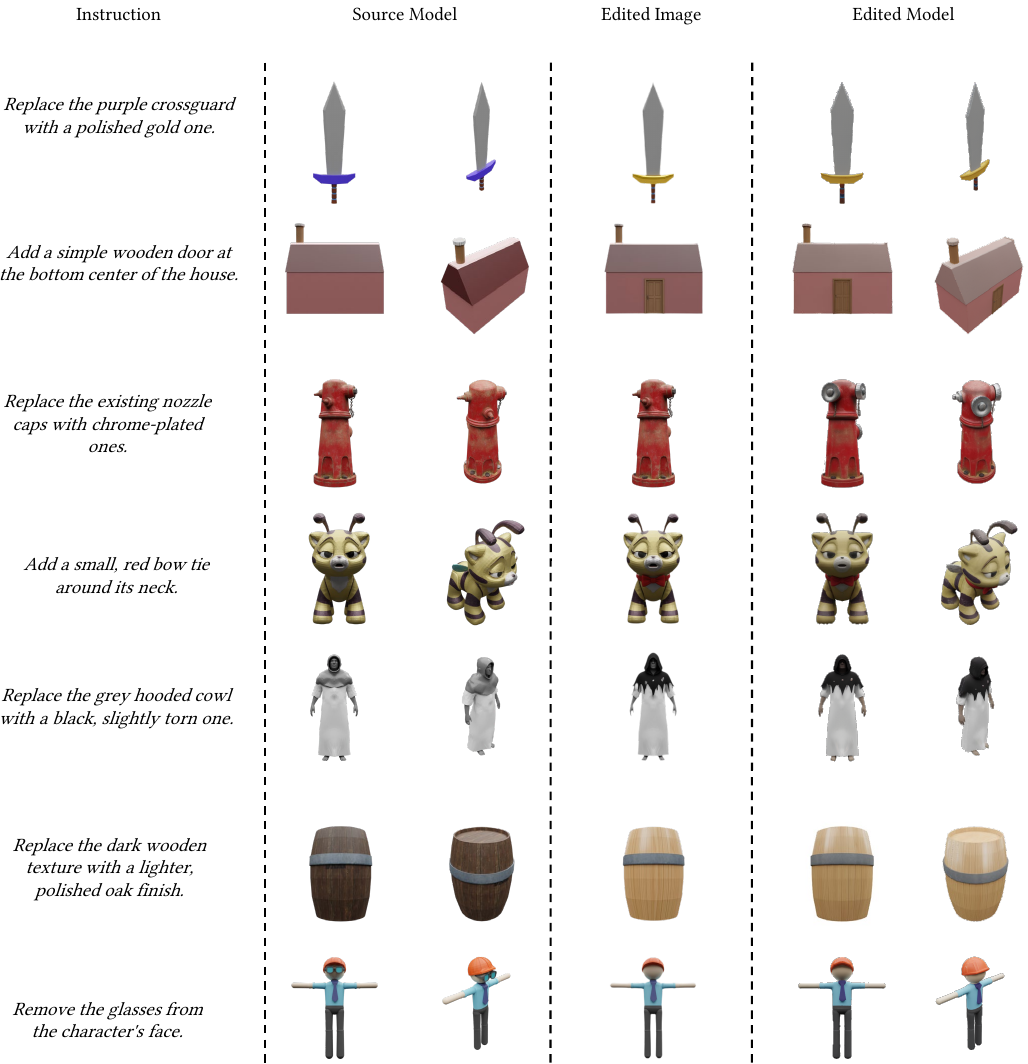}
    \caption{\textbf{More results.}}
    \label{fig:more_results}
\end{figure*}

\section{More Results}
\label{sec:supp_more_results}

Figure~\ref{fig:more_results} presents additional qualitative results across diverse editing operations, including part addition, removal, and material replacement. Our method successfully executes precise local edits while preserving the unedited regions. Furthermore, we highly encourage readers to refer to the \textbf{attached supplementary webpage}, which provides $360^\circ$ multi-view rendered videos to better demonstrate the structural integrity and cross-view consistency of our edited 3D assets.

\section{Comparison with VoxHammer}
\begin{figure*}[t]
    \centering
    \includegraphics[width=0.9\linewidth]{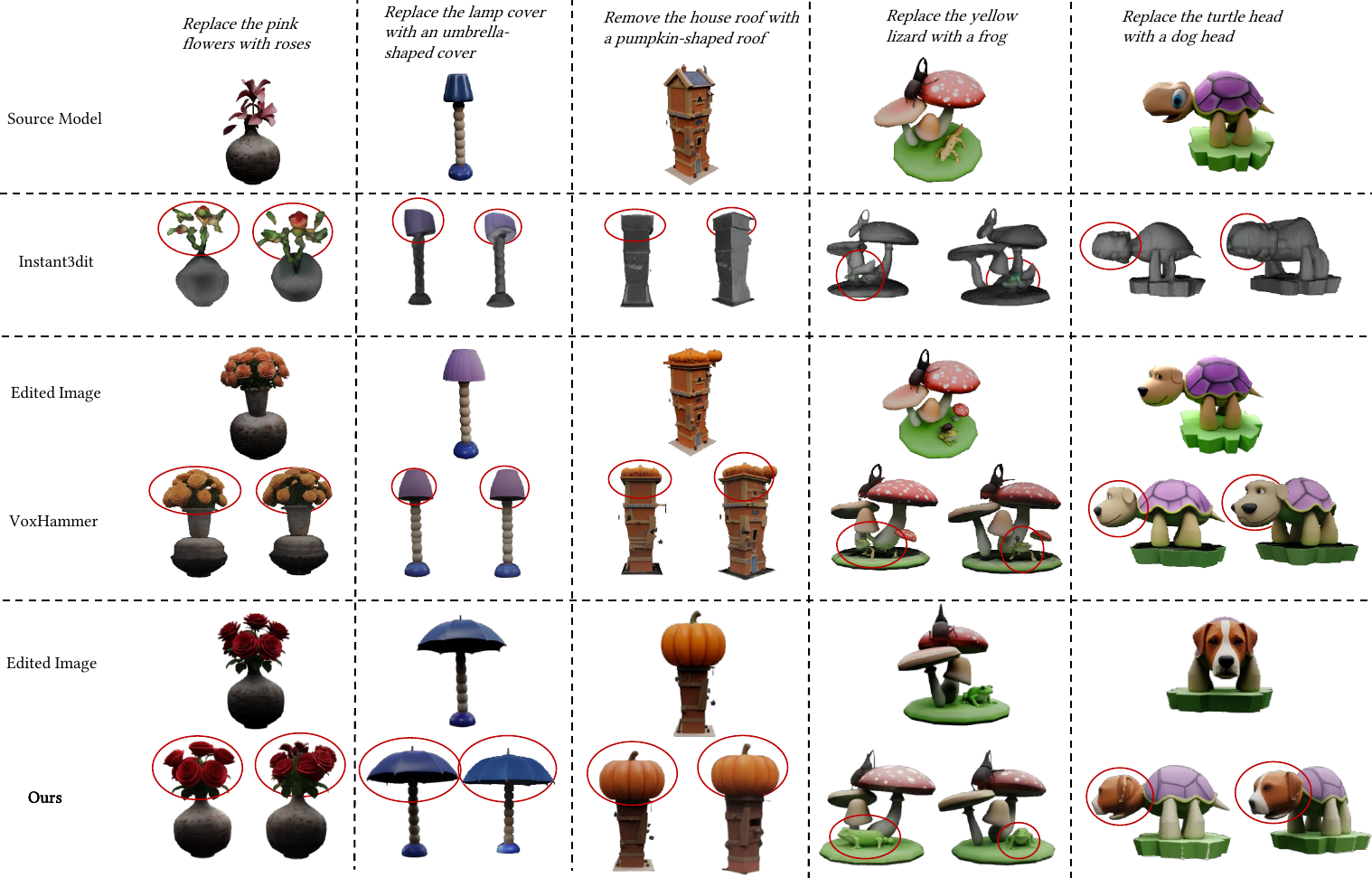}
    \caption{\textbf{Qualitative results with samples in Edit3D-Bench.}}
    \label{fig:edit3dbench}
\end{figure*}
To compare our method with VoxHammer, we evaluate our method on samples from Edit3D-Bench. Figure~\ref{fig:edit3dbench} presents qualitative results on representative samples from Edit3D-Bench. Since our method is designed to operate under view-conditioned image editing inputs rather than the original benchmark setting, we render a novel view for each sample and regenerate the edited image from that view as the input to our pipeline.  The results show that our method achieves competitive editing quality while better preserving structural consistency across views, without requiring manually annotated 3D masks.
\label{sec:edit_3d_bench}

\begin{figure*}[t]
    \centering
    \hfill
    \begin{minipage}{0.48\linewidth}
        \centering
        \includegraphics[width=\linewidth]{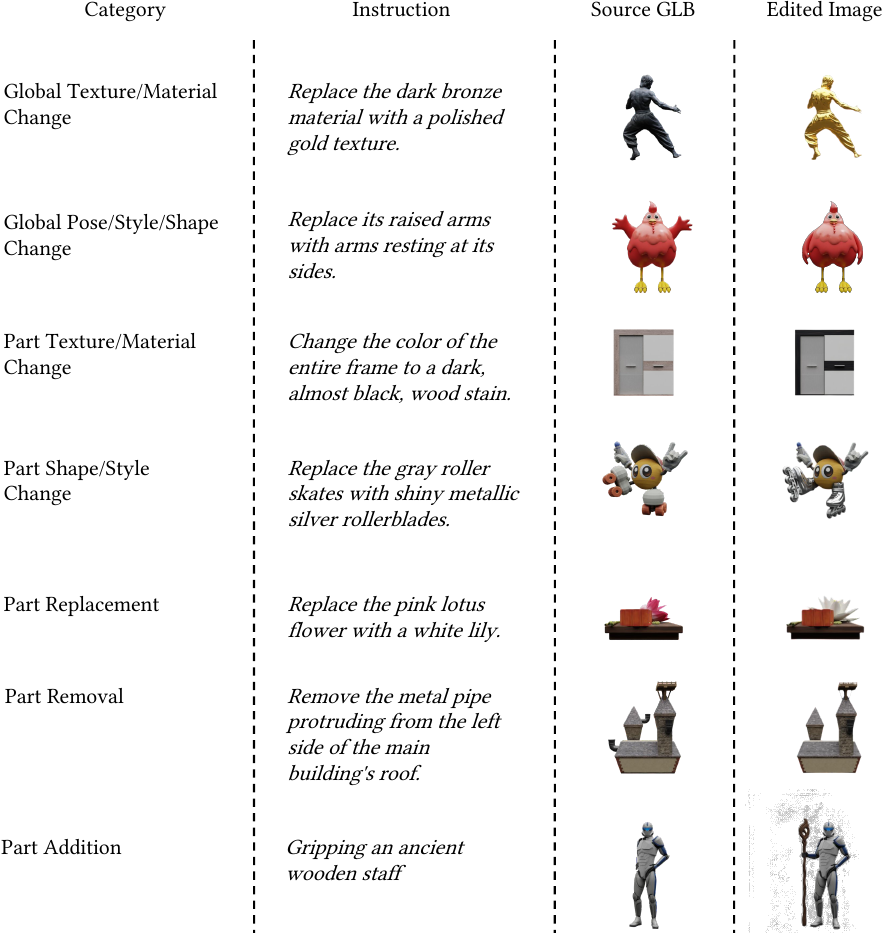}
        \caption{Representative samples from our 3D editing dataset. We showcase various editing types including texture alterations, local geometric edits, and broad stylistic modifications. Each example displays the source asset, the corresponding instruction, and the targeted outcome.}
        \label{fig:supp_dataset_samples}
    \end{minipage}
    \hfill
    \begin{minipage}{0.48\linewidth}
        \centering
        \includegraphics[width=\linewidth]{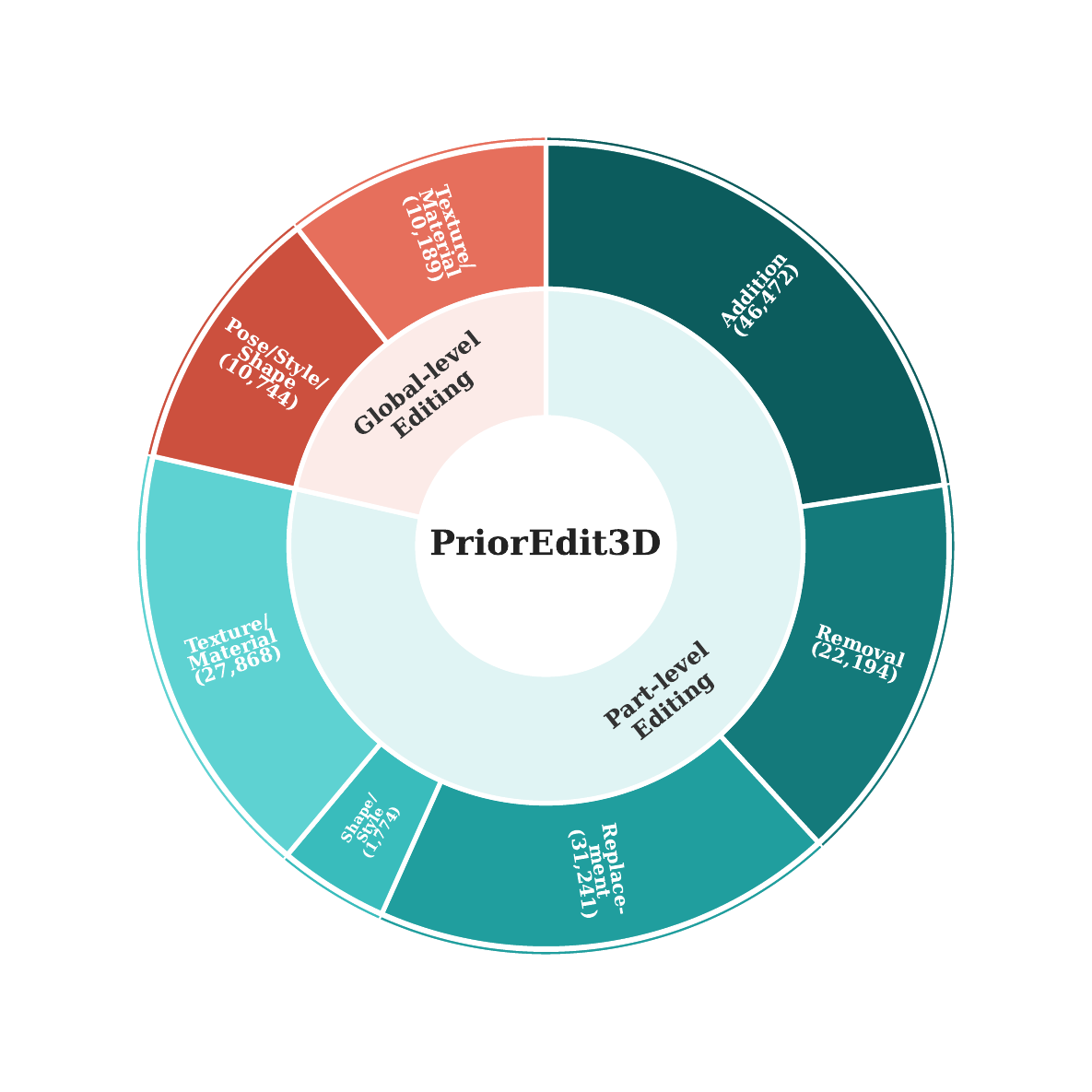}
        \caption{The proportion of each edit type in our dataset.}
        \label{fig:supp_dataset_pie}
    \end{minipage}
    \hfill
\end{figure*}

\section{Dataset Statistics}
\label{sec:supp_dataset}
To evaluate our model across diverse scenarios, our dataset encompasses a broad spectrum of 3D object modifications. We categorize the editing instructions into several primary types based on the nature of the edits:
\begin{itemize}
    
    \item Part-level Editing
    \begin{itemize}
        \item Addition(46472): Add a part to the object (e.g., ``Add a window to the wall'', ``Add a plate on the table'')
        \item Removal(22194): Remove a certain part from the object (e.g., ``Remove the glasses from the boy'', ``Remove the liquid from the cup'')
        \item Replacement(31241): Replace a part into another one (e.g., ``Replace the button on the panel with a lever'', ``Replace the trouser with a skirt'')
        \item Shape / Style Change(1774): Changing the geometry of a specific part (e.g., ``Make the button on the panel into a round one'', ``Make the girl's hair curly'')
        \item Texture / Material Change(27868): Changing the appearance of a specific part without altering its geometry (e.g, ``Make the shirt red'', ``Replace the wooden handle with an iron one'')
    \end{itemize}
    \item Global-level Editing
    \begin{itemize}
        \item Pose / Style / Shape Change(10744): Changing the global geometry of an object (e.g., ``make it spiky'', ``make the boy wave his hand'')
        \item Texture / Material Change(10189):  Changing the global appearance of an object without altering its geometry (e.g., ``make it blue'', ``use a wooden texture'').
        
    \end{itemize}
\end{itemize}

The proportion of each type is shown in Figure ~\ref{fig:supp_dataset_pie}.

Detailed sample visualizations of these editing types, along with their instructions,  source 3D asset render results and expected outcomes, are illustrated in Fig.~\ref{fig:supp_dataset_samples}.

\section{Detailed Experimental Configuration}
\label{sec:supp_experimental_configuration}

Table~\ref{tab:supp_experimental_configuration} summarizes the model architecture and all tuned training parameters used in our experiments.

\begin{table*}[t]
    \centering
    \caption{Detailed model and training configuration used in our experiments.}
    \label{tab:supp_experimental_configuration}
    \scriptsize
    \setlength{\tabcolsep}{4pt}
    \renewcommand{\arraystretch}{1.12}
    \begin{tabularx}{0.84\textwidth}{@{}p{0.15\textwidth}p{0.20\textwidth}X@{}}
        \csname toprule\endcsname
        {\bfseries Component} & \textbf{Parameter} & \textbf{Value} \\
        \midrule
        \multicolumn{3}{l}{\textbf{Model architecture}} \\
        Denoiser & Model & SparseStructureFlowModel \\
        Auxiliary model & Model & SparseStructureFlowModel; architecture identical to the denoiser \\
        Both  & Sparse resolution; patch size & 16; 1 \\
        Both  & Input/output channels & 32 / 32 \\
        Both & Model/conditioning channels & 1280 / 1280 \\
        Both & Blocks; attention heads; MLP ratio & 36; 32; 4 \\
        Both & Positional encoding; QK normalization & Absolute positional encoding (APE); QK RMS normalization enabled \\
        Both & Numerical precision & bfloat16 \\
        \midrule
        \multicolumn{3}{l}{\textbf{Data, conditioning, and initialization}} \\
        Dataset & Dataset implementation & ObjaverseUniLat \\
        Image conditioning & Encoder & dinov3\_vith16plus \\
        \midrule
        \multicolumn{3}{l}{\textbf{Optimization and DMD}} \\
        Training & Maximum steps; checkpoint interval; batch size & 18,000; every 2,000 steps; 1 per GPU \\
        Warmup & Warmup steps & 500 \\
        Optimizer & Type & AdamW \\
        Optimizer & Learning rate; weight decay; betas & $1\times10^{-5}$; 0; $(0.0, 0.9)$ \\
        DMD & Enabled; weight & Yes; 0.01 \\
        Auxiliary model & Updates per editor step & 10 \\
        \midrule
        \multicolumn{3}{l}{\textbf{Supervision losses}} \\
        Pixelwise loss & Type; weight & $\ell_1$; 1.0 \\
        LPIPS loss & Weight & 0.2 \\
        VLM loss & Backbone; weight & Qwen3-VL-4B-Instruct; 0.0005 \\
        \midrule
        \multicolumn{3}{l}{\textbf{Sampling configuration}} \\
        Reverse denoising & Timestep sequence & 20 descending levels from 1.00 to 0.05 with an interval of 0.05 \\
        \bottomrule
    \end{tabularx}
\end{table*}

\section{Training}
\label{sec:training}

The training procedure of our method is summarized in Algorithms~\ref{alg:2d_supervision}, \ref{alg:unroll}, and \ref{alg:training}. Specifically, Algorithm~\ref{alg:2d_supervision} presents the 2D supervision obtained through differentiable rendering, Algorithm~\ref{alg:unroll} describes the short reverse unroll used to predict the edited latent, and Algorithm~\ref{alg:training} summarizes the overall DMD-regularized alternating optimization process.

\begin{algorithm}[H]

\caption{2D Supervision via Differentiable Rendering}
\label{alg:2d_supervision}
\begin{algorithmic}[1]
\Require Source latent $x_{\mathrm{src}}$, edited latent $\hat{x}_0$, target image $I_{\mathrm{tgt}}^{\pi_0}$, instruction $d$
\Require Differentiable renderer $R$, frozen VLM, optional mask $M^{\pi_0}$
\Require Weights $\lambda_{\mathrm{fg}}, \lambda_{\mathrm{lpips}}$

\State $(I_{\mathrm{pred}}^{\pi_0}, O_{\mathrm{pred}}^{\pi_0}) \gets R(\hat{x}_0; \pi_0)$
\If{$M^{\pi_0}$ is available}
    \State $\mathcal{L}_{\mathrm{pixel}} \gets \|M^{\pi_0}\odot(I_{\mathrm{pred}}^{\pi_0}-I_{\mathrm{tgt}}^{\pi_0})\|_1$
    \Statex \hspace{\algorithmicindent}$\qquad\qquad\quad + \lambda_{\mathrm{fg}}\|(1-M^{\pi_0})\odot O_{\mathrm{pred}}^{\pi_0}\|_1$
    \Statex \hspace{\algorithmicindent}$\qquad\qquad\quad + \lambda_{\mathrm{lpips}}\,\mathrm{LPIPS}(I_{\mathrm{pred}}^{\pi_0}, I_{\mathrm{tgt}}^{\pi_0})$
\Else
    \State $\mathcal{L}_{\mathrm{pixel}} \gets \|I_{\mathrm{pred}}^{\pi_0}-I_{\mathrm{tgt}}^{\pi_0}\|_1
    + \lambda_{\mathrm{lpips}}\,\mathrm{LPIPS}(I_{\mathrm{pred}}^{\pi_0}, I_{\mathrm{tgt}}^{\pi_0})$
\EndIf

\State $I_{\mathrm{src}}^{\pi_1} \gets \mathrm{RGB}(R(x_{\mathrm{src}};\pi_1))$, \quad $I_{\mathrm{pred}}^{\pi_1} \gets \mathrm{RGB}(R(\hat{x}_0;\pi_1))$
\State $I_{\mathrm{src}}^{\pi_2} \gets \mathrm{RGB}(R(x_{\mathrm{src}};\pi_2))$, \quad $I_{\mathrm{pred}}^{\pi_2} \gets \mathrm{RGB}(R(\hat{x}_0;\pi_2))$

\State $(o^{\mathrm{IF}}_{\mathrm{Yes}},o^{\mathrm{IF}}_{\mathrm{No}}) \gets \mathrm{VLMlogits}(I_{\mathrm{src}}^{\pi_1}, I_{\mathrm{pred}}^{\pi_1}, P_{\mathrm{IF}}(d))$
\State $(o^{\mathrm{IP}}_{\mathrm{Yes}},o^{\mathrm{IP}}_{\mathrm{No}}) \gets \mathrm{VLMlogits}(I_{\mathrm{src}}^{\pi_2}, I_{\mathrm{pred}}^{\pi_2}, P_{\mathrm{IP}})$

\State $\mathcal{L}_{\mathrm{vlm}} \gets -\log \sigma(o^{\mathrm{IF}}_{\mathrm{Yes}}-o^{\mathrm{IF}}_{\mathrm{No}})
-\log \sigma(o^{\mathrm{IP}}_{\mathrm{Yes}}-o^{\mathrm{IP}}_{\mathrm{No}})$

\State \Return $\mathcal{L}_{\mathrm{pixel}}, \mathcal{L}_{\mathrm{vlm}}$
\end{algorithmic}
\end{algorithm}

\begin{algorithm}[H]
\caption{Short Reverse Unroll for Edited Latent Prediction}
\label{alg:unroll}
\begin{algorithmic}[1]
\Require Source latent $x_{\mathrm{src}}$, condition image $c$, student velocity field $v_\theta$
\Require Timesteps $\{t_i\}_{i=0}^{N}$ with $t_N>\cdots>t_1>t_0=0$
\Require Exit range $[s_{\min}, s_{\max}]$
\State Sample $\epsilon \sim \mathcal{N}(0,I)$ and $s \sim \mathcal{U}\{s_{\min},\ldots,s_{\max}\}$
\State $\hat{x}_{t_N} \gets \epsilon$
\For{$i=N$ down to $s+1$}
    \State $\widetilde{x}_{t_i} \gets \operatorname{Concat}_{\mathrm{token}}(\hat{x}_{t_i},x_{\mathrm{src}})$
    \State $\hat{x}_{t_{i-1}} \gets \hat{x}_{t_i} - (t_i-t_{i-1})\,v_\theta(\widetilde{x}_{t_i}, t_i, c)$
    \Comment{stop gradient}
\EndFor
\State $\widetilde{x}_{t_s} \gets \operatorname{Concat}_{\mathrm{token}}(\hat{x}_{t_s},x_{\mathrm{src}})$
\State $\hat{x}_0 \gets \hat{x}_{t_s} - t_s\,v_\theta(\widetilde{x}_{t_s}, t_s, c)$
\Comment{backpropagate only through the final jump}
\State \Return $\hat{x}_0$
\end{algorithmic}
\end{algorithm}

\begin{algorithm}[H]
\caption{DMD-Regularized Alternating Training of PriorEdit3D}
\label{alg:training}
\begin{algorithmic}[1]
\Require Student editor $G_\theta$, fake model $F_\varphi$, prior teacher $G_{\mathrm{real}}$
\Require Warm-up steps $W$, fake-model updates per editor step $K$
\Require Weights $\lambda_{\mathrm{vlm}}, \lambda_{\mathrm{DMD}}$
\Require Batch $\mathcal{B}=\{x_{\mathrm{src}}, I_{\mathrm{src}}^{\pi_0}, I_{\mathrm{tgt}}^{\pi_0}, d, M^{\pi_0}(\text{optional})\}$

\While{training}
    \If{$\mathrm{global\_step} < W$}
        \Comment{identity warm-up}
        \State $\hat{x}_0 \gets \Call{ShortReverseUnroll}{x_{\mathrm{src}}, I_{\mathrm{src}}^{\pi_0}}$
        \State $\mathcal{L}_{\mathrm{id}} \gets \|\hat{x}_0-x_{\mathrm{src}}\|_2^2$
        \State Update $\theta$ using $\mathcal{L}_{\mathrm{id}}$
        \State \textbf{continue}
    \EndIf

    \For{$k=1$ to $K$}
        \Comment{update fake model}
        \State $\hat{x}_0 \gets \Call{ShortReverseUnroll}{x_{\mathrm{src}}, I_{\mathrm{tgt}}^{\pi_0}}$ \Comment{no grad to $\theta$}
        \State Sample $t\sim \mathcal{U}(0,1)$ and $\epsilon \sim \mathcal{N}(0,I)$
        \State $\hat{x}_t \gets (1-t)\hat{x}_0 + t\epsilon$
        \State $v \gets \epsilon-\hat{x}_0$
        \State $\mathcal{L}_{\mathrm{fake}} \gets \|F_\varphi(\hat{x}_t,t,I_{\mathrm{tgt}}^{\pi_0})-v\|_2^2$
        \State Update $\varphi$
    \EndFor

    \Comment{update editor}
    \State $\hat{x}_0 \gets \Call{ShortReverseUnroll}{x_{\mathrm{src}}, I_{\mathrm{tgt}}^{\pi_0}}$
    \State $(\mathcal{L}_{\mathrm{pixel}}, \mathcal{L}_{\mathrm{vlm}}) \gets
    \Call{TwoDSupervision}{x_{\mathrm{src}}, \hat{x}_0, I_{\mathrm{tgt}}^{\pi_0}, d, M^{\pi_0}}$

    \State Sample $t\sim \mathcal{U}(0,1)$ and $\epsilon \sim \mathcal{N}(0,I)$
    \State $\hat{x}_t \gets (1-t)\hat{x}_0 + t\epsilon$
    \State $v_{\mathrm{real}} \gets G_{\mathrm{real}}(\hat{x}_t,t,I_{\mathrm{tgt}}^{\pi_0})$
    \State $v_{\mathrm{fake}} \gets F_\varphi(\hat{x}_t,t,I_{\mathrm{tgt}}^{\pi_0})$
    \State Compute $\mathcal{L}_{\mathrm{DMD}}$ from Eq.~(7)

    \State $\mathcal{L} \gets \mathcal{L}_{\mathrm{pixel}} + \lambda_{\mathrm{vlm}}\mathcal{L}_{\mathrm{vlm}} + \lambda_{\mathrm{DMD}}\mathcal{L}_{\mathrm{DMD}}$
    \State Update $\theta$
\EndWhile
\end{algorithmic}
\end{algorithm}

\section{Training Cost}
\label{sec:supp_training_cost}

We report the training-time computational cost of different supervision settings in Table~\ref{tab:supp_training_cost}.
For DMD-based settings, the reported time is measured over one complete student-update cycle, consisting of one student update and 10 fake-model updates.
The ``Fake Model'' row reports the cost of a single fake-model update.

\begin{table}[t]
    \centering
    \caption{
    Training-time computational cost and peak GPU memory under different supervision settings.
    }
    \label{tab:supp_training_cost}
    \small
    \setlength{\tabcolsep}{12pt}
    \renewcommand{\arraystretch}{1.12}
    \resizebox{\linewidth}{!}{
    \begin{tabular}{lccc}
        \toprule
        \textbf{Setting} & \textbf{Forward} & \textbf{Backward} & \textbf{Peak Memory} \\
        \midrule
        Fake Model (UniLat3D) & 6.498 s & 1.061 s & -- \\
        Pixel & 7.782 s & 1.246 s & 38 GB \\
        Pixel + DMD & 73.006 s & 11.730 s & 43 GB \\
        Pixel + DMD + VLM & 73.252 s & 11.878 s & 59 GB \\
        Inference & 7 s & -- & 13 GB \\
        \bottomrule
    \end{tabular}
    }
\end{table}

The additional training overhead mainly comes from the repeated fake-model updates required by DMD.
In comparison, introducing VLM supervision adds only a small amount of computation but increases peak GPU memory from 43 GB to 59 GB.
These additional costs are incurred only during training; the VLM, fake model, and differentiable supervision pipeline are removed at inference time, and PriorEdit3D remains a feed-forward editor with an inference time of approximately 7 seconds per asset.

\end{document}